\pdfoutput=1

\documentclass[11pt]{article}

\usepackage[]{EMNLP2023}

\usepackage{times}
\usepackage{latexsym}

\usepackage[T1]{fontenc}

\usepackage[utf8]{inputenc}

\usepackage{microtype}

\usepackage{inconsolata}

\usepackage{graphicx}

\usepackage{graphicx}               
\usepackage{tabularx}               
\usepackage{booktabs}
\usepackage{amsmath}
\usepackage{stmaryrd}
\usepackage{xcolor}
\usepackage{soul}

\newcolumntype{C}{>{\centering\arraybackslash}X}
\usepackage{multirow}               
\usepackage{diagbox}                
\usepackage{hhline}                 
\usepackage{color}                  
\usepackage{amsmath}                
\usepackage{mathtools}              
\usepackage{enumitem}               

\usepackage{subfigure}
\usepackage{booktabs}
\usepackage{extarrows}
\usepackage{makecell}

\usepackage{soul}

\definecolor{RoseQuartzBg}{HTML}{F7CAC9}
\definecolor{RoseQuartz}{HTML}{F5A798}
\definecolor{Serenity}{HTML}{92A8D1}
\definecolor{OrangeRed}{rgb}{1.0, 0.27, 0.0}
\definecolor{Red}{rgb}{1.0, 0.0, 0.0}
\definecolor{Turquoise}{HTML}{0F4C81}

\usepackage{xparse}
\NewDocumentCommand{\lifu}{ mO{} }{\textcolor{OrangeRed}{\textsuperscript{\textit{Lifu}}\textsf{\textbf{\small[#1]}}}}
\NewDocumentCommand{\minqian}{ mO{} }{\textcolor{violet}{\textsuperscript{\textit{Minqian}}\textsf{\textbf{\small[#1]}}}}
\NewDocumentCommand{\zhiyang}{ mO{} }{\textcolor{Serenity}{\textsuperscript{\textit{Zhiyang}}\textsf{\textbf{\small[#1]}}}}
\NewDocumentCommand{\ying}{ mO{} }{\textcolor{teal}{\textsuperscript{\textit{Ying}}\textsf{\textbf{\small[#1]}}}}

\NewDocumentCommand{\jy}{ mO{} }{\textcolor{brown}{\textsuperscript{\textit{jy}}\textsf{\textbf{\small[#1]}}}}

\usepackage{xcolor}
\usepackage[linesnumbered,ruled,vlined]{algorithm2e}

\SetCommentSty{mycommfont}

\newcommand{\method}{\textsc{Socratic Questioning}}

\newcommand{\selfQuestion}{\textsc{Self-Questioning}}

\newcommand{\FQG}{FQG}

\newcommand{\FQA}{FQA}
\newcommand{\QAhint}{QA-to-Hint}
\newcommand{\QAtohint}{QA2H}

\newcommand{\VQG}{VQG}
\newcommand{\VQA}{VQA}
\newcommand{\BLIP}{BLIP-2}

\definecolor{fig_blue}{HTML}{9DCEC9}
\definecolor{fig_red}{HTML}{EA8575}
\definecolor{fig_yellow}{HTML}{F5C084}
\definecolor{back_green}{HTML}{43AA8B}
\definecolor{down_pink}{HTML}{EE7674}
\definecolor{back_pink}{HTML}{EE7674}
\definecolor{visual_orange}{HTML}{f5c083}
\definecolor{fact_pink}{HTML}{eb8575}

\newcommand{\visualPercept}{Visual Perception}

\title{\includegraphics[scale=0.02]{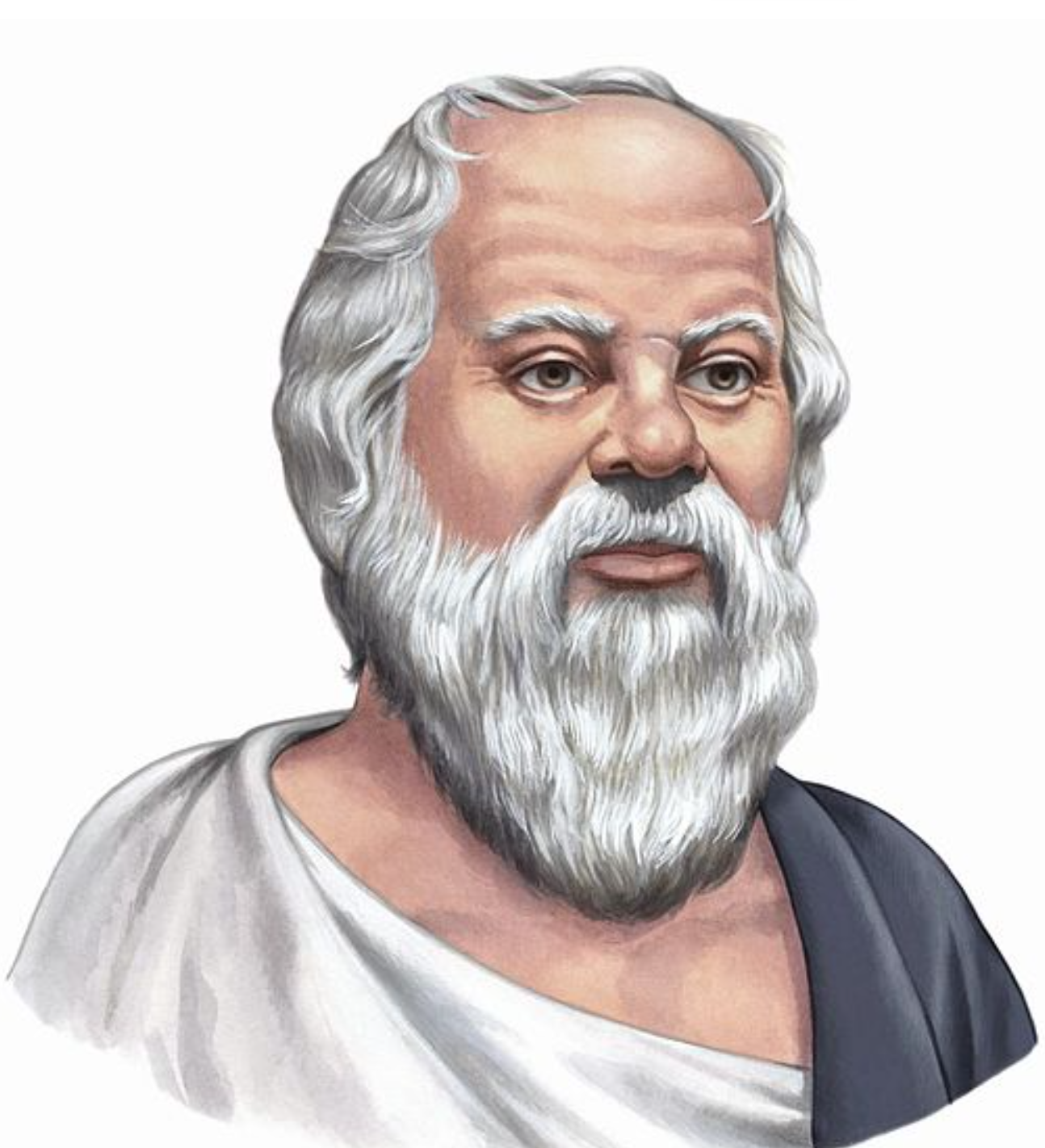} The Art of \method{}: Recursive Thinking with Large Language Models}

\author{Jingyuan Qi$^{*\spadesuit}$ \quad Zhiyang Xu$^{*\spadesuit}$ \quad Ying Shen$^{\dagger\spadesuit}$ \quad Minqian Liu$^{\dagger\spadesuit}$ \\ \textbf{Di Jin}$^\heartsuit$ \quad \textbf{Qifan Wang}$^\clubsuit$ \quad \textbf{Lifu Huang}$^{\spadesuit}$ \\
  $^{\spadesuit}$Virginia Tech \quad $^\heartsuit$Amazon Inc. \quad $^\clubsuit$Meta AI \\
  \texttt{\{jingyq1,zhiyangx,yings,minqianliu,lifuh\}@vt.edu} \\ {djinamzn@amazon.com} \quad {wqfcr@meta.com} \\}
  
\begin{document}
\maketitle

\begin{abstract}

Chain-of-Thought (CoT) prompting enables large language models to solve complex reasoning problems by generating intermediate steps.
However, confined by its inherent single-pass and sequential generation process, CoT heavily relies on the initial decisions, causing errors in early steps to accumulate and impact the final answers. In contrast, humans adopt recursive thinking when tackling complex reasoning problems, i.e., iteratively breaking the original problem into approachable sub-problems and aggregating their answers to resolve the original one. 
Inspired by the human cognitive process, we propose \method{}, a divide-and-conquer style algorithm that mimics the recursive thinking process. 
Specifically, \method{} leverages large language models to raise and answer sub-questions until collecting enough information to tackle the original question. 
Unlike CoT, \method{} explicitly navigates the thinking space, stimulates effective recursive thinking, and is more robust towards errors in the thinking process.
Extensive experiments on several complex reasoning tasks, including MMLU, MATH, LogiQA, and visual question-answering demonstrate 
significant performance improvements over the state-of-the-art prompting methods, such as CoT, and Tree-of-Thought. The qualitative analysis clearly shows that the intermediate reasoning steps elicited by \method{} are similar to humans' recursively thinking process of complex reasoning problems\footnote{* Co-first Authors, $\dagger$ Co-second Authors}\footnote{All the programs and necessary resources are released in \url{https://github.com/VT-NLP/SOCRATIC-QUESTIONING}}. 

\end{abstract}

\section{Introduction}

\begin{figure}[ht!]
    \centering
    \includegraphics[width=0.45\textwidth]{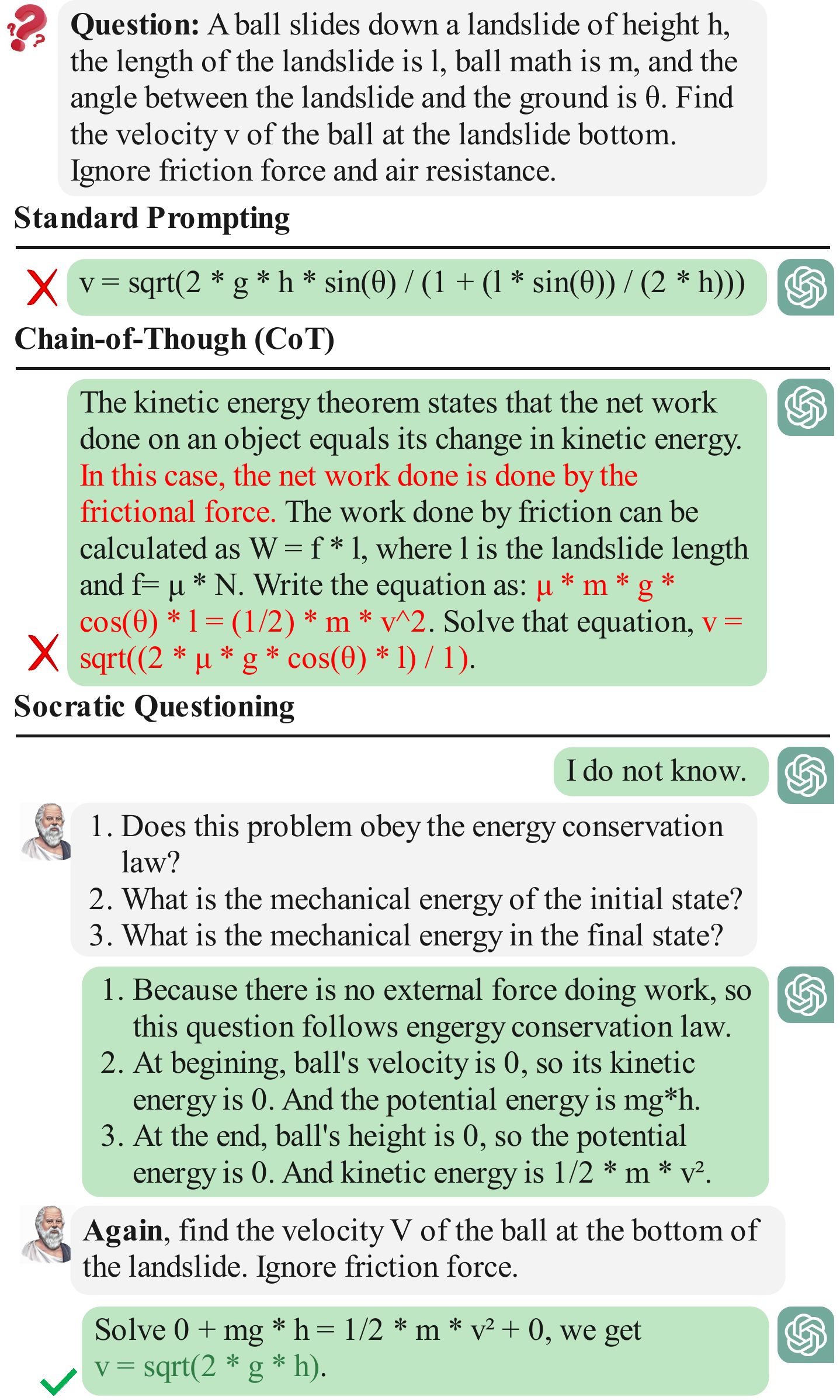}
    \vspace{-2mm}
    \caption{Example of a complex question solved by the Standard Prompting, Chain-of-Thought, and \method{}. Accumulated incorrect reasoning are highlighted in \textcolor{red}{red}.
    }
    \label{fig:question_exp}
\end{figure}

\begin{quote}
\textit{The art of Socratic Questioning is important for critical thinkers and excellence of thought. What Socratic adds is systematicity, depth, and a keen interest in assessing the plausibility of things.} 
\\
\centerline{- L. ELDER and R. PAUL, 1998}   
\end{quote}

One unique capability that allows humans to excel at solving complex reasoning problems is \textit{recursive thinking}. If the answer is not immediately achievable, humans think deeper by recursively decomposing the complex problem into simpler and solvable sub-problems.
\begin{figure*}[!tbh]
    \centering
    \includegraphics[width=0.9\textwidth]{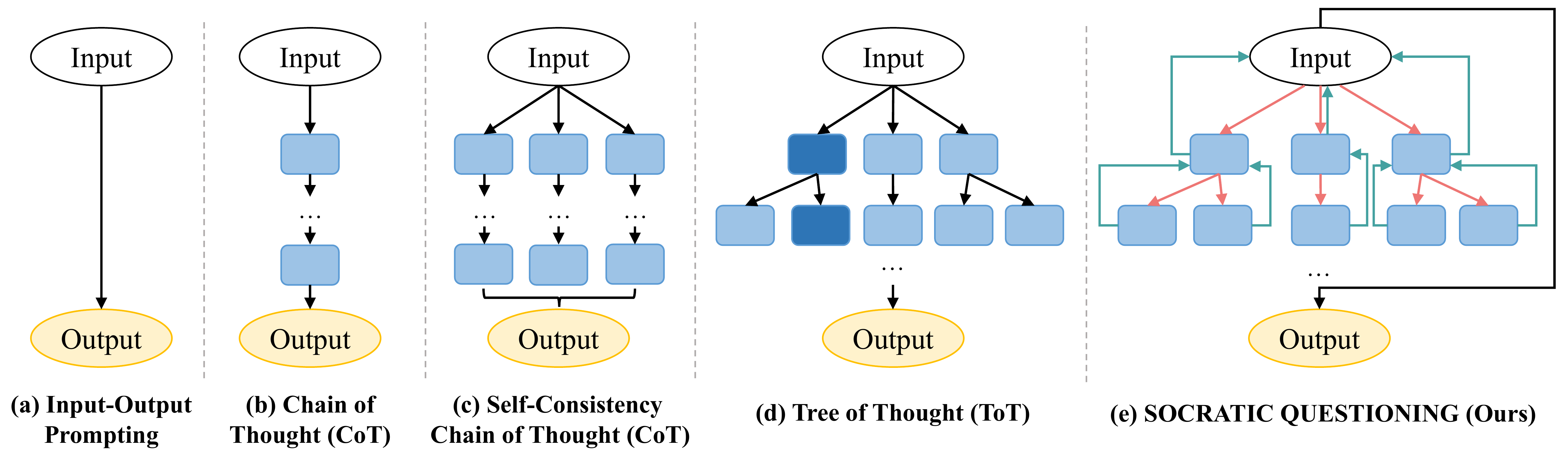}
    \vspace{-2mm}
    \caption{\textbf{Schematic comparison of various prompting methods.} Each blue rectangle box represents a \textit{thought} serving as an intermediate reasoning step in the problem-solving process. \method{} incorporates both a \textcolor{down_pink}{top-down exploration} process (in \textcolor{down_pink}{red} line) to deconstruct complex problems into smaller sub-questions and a \textcolor{back_green}{bottom-up backtracking} process (in \textcolor{back_green}{green} line) to recursively solve these sub-questions and gather solutions for higher-level problems.
    }
    \vspace{-5mm}
    \label{fig:comparison}
\end{figure*}

Recently, by scaling up the parameters, large-scale language models (LLMs)~\cite{brown2020language, Chung2022FlanT5, ChatGPT, Touvron2023LLAMA} gain emerging capabilities, such as Chain-of-Thought (CoT)~\cite{wei2022COT} which decomposes the complex problem and solves it step by step. Though CoT has been proven to be effective on various complex reasoning tasks, it's in nature a single-pass and sequential thinking process that generates the next step based on previous steps, thus only exploring a single way of thinking to approach a problem and easily accumulating errors from previous steps~\cite{Turpin2023CoTUnfaithful}. In addition, CoT lacks the ability to 
refine the already generated reasoning path, as shown in Figure~\ref{fig:question_exp}. 

Inspired by the recursive thinking of humans, we propose \method{}, a novel divide-and-conquer fashion algorithm that prompts language models to solve complex reasoning problems.
As shown in Figure~\ref{fig:comparison} (e), \method{} consists of a \textcolor{down_pink}{top-down} exploration process and a \textcolor{back_green}{bottom-up} backtracking process. Specifically, in the top-down exploration process, the original complex problem is decomposed into simpler or related sub-problems until the sub-problems can be solved. In the bottom-up backtracking process, the solutions to the sub-problems are returned and selectively used to solve the original problem.
The fundamental component that drives \method{} is a \selfQuestion{} (SQ) module, that leverages large-scale language models to proactively raise and answer questions that are essential to solving the target question. 
\method{} recursively backtracks and tailors the intermediate thoughts acquired from \selfQuestion{} until reaching an answer to the original input question. It explicitly navigates the thinking space and is more robust towards thinking errors compared with previous prompting methods including CoT, Self-Consistency Chain-of-Thought~\cite{wang2022self}, and Tree-of-Thought~\cite{yao2023ToT}, as shown in Figure~\ref{fig:comparison}.

To show the effectiveness of \method{}, we conduct extensive experiments on various complex reasoning tasks including the chemistry and physics tasks 
~\cite{Hendrycks2020multitask}, mathematical tasks
~\cite{Hendrycks2021math}, and reading comprehension tasks
\cite{Liu2020logiqa}. Additionally, we showcase the generalizability of our method by conducting experiments with few-shot multimodal reasoning on VQA-V2~\cite{goyal2017making}, OK-VQA~\cite{okvqa}, and AOK-VQA~\cite{schwenk2022okvqa} datasets.
Experimental results indicate that
\method{} substantially improves performance over CoT, SC-CoT, and ToT across all language tasks and outperforms several strong baselines in few-shot multimodal reasoning. The qualitative analysis further demonstrates that \method{} is capable of eliciting the intermediate reasoning steps through \selfQuestion{}, like a critical thinker, and solving complex reasoning problems.
The main contributions of our paper are as follows:

\begin{itemize}
\itemsep -0.5ex
    \item We propose \method{}, a novel prompting algorithm that can navigate the cognitive thinking space in a \textbf{\textit{recursive}} manner.
    \item We introduce the \selfQuestion{} module, a core component that actively probes complex problems from various perspectives by raising and addressing questions essential for solving the main problem. 
    \item Our approach achieves significant improvements over the previous prompting methods in various complex reasoning tasks.
\end{itemize}
\section{Related Work}

\begin{figure*}[!tbh]
    \centering
    \includegraphics[width=0.9\textwidth]{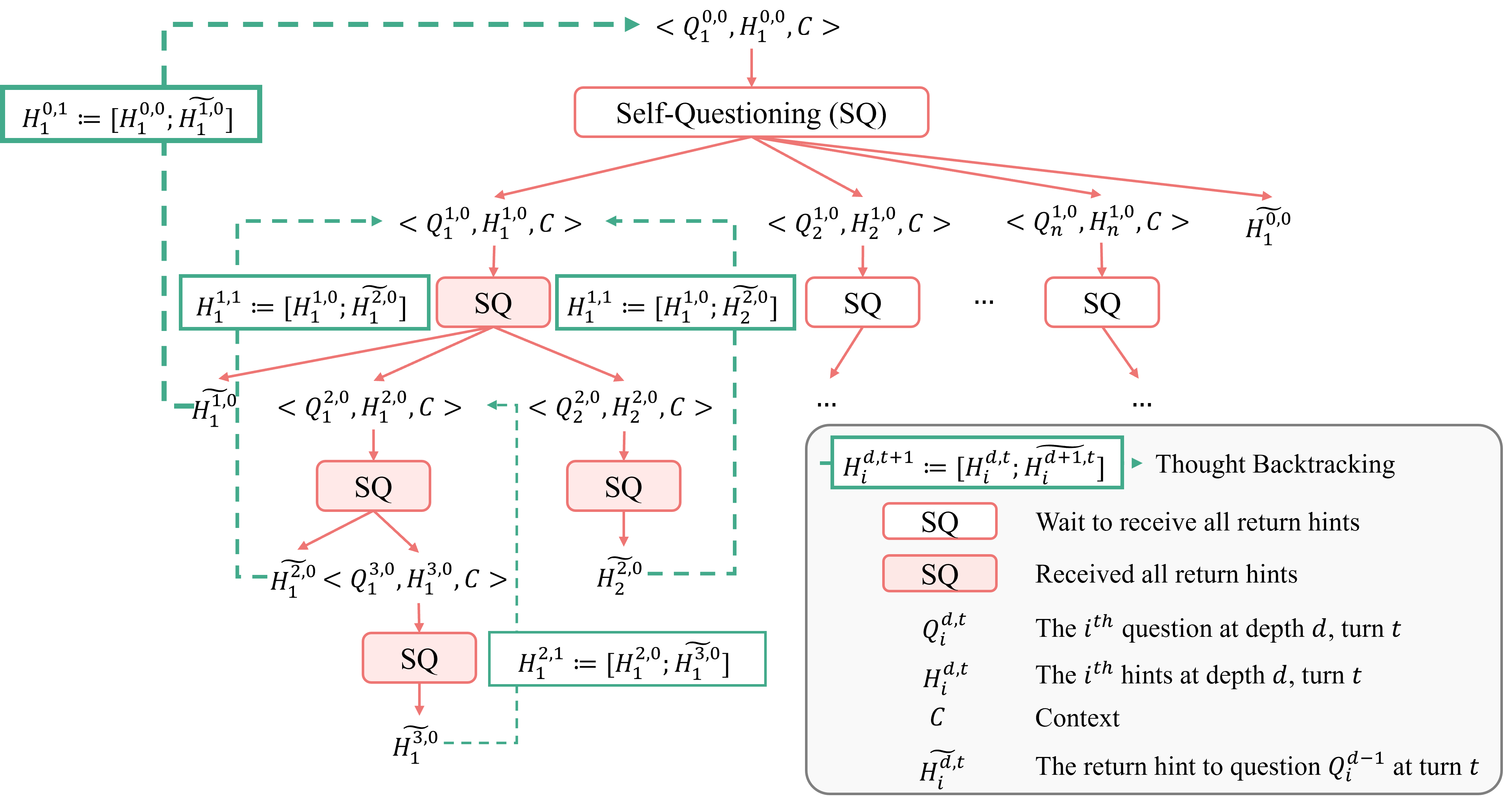}
    \caption{Overview of our \method{} algorithm.
    }
    \vspace{-3mm}
    \label{fig:overview}
\end{figure*}

\paragraph{Prompting Large Language Models} With the scaling of both modal size and corpus size, large language models (LLMs) such as GPT-3 \cite{brown2020language} and ChatGPT \cite{ChatGPT} have exhibited emergent abilities, including prompting \cite{brown2020language}, in-context learning~\cite{dong2022survey}, and commonsense reasoning \cite{weiemergent}. One notable example of emergent  abilities is the Chain-of-Thought (CoT) \cite{wei2022COT} which steers large language models to resolve complex problems by guiding them to produce a sequence of intermediate steps before giving the final answer. Self-Consistency Chain-of-Thought (SC-CoT)~\cite{wang2022self} improves naive CoT by sampling multiple reasoning paths and selecting the most consistent answer. SC-CoT is based on the assumption that given a complex reasoning problem, multiple reasoning paths can lead to the unique correct answer. Tree-of-Thought (ToT)~\cite{yao2023ToT} proposes to break the thinking process into small steps and at each step, the language model deliberately decides a set of next steps to try.




\paragraph{Multimodal Reasoning with Large Language Models} 
Recent studies have explored the collaboration among diverse language and visual models \cite{Yang2022PICa,Zeng2022SocraticModels,Huang2022planningLM}. For example, PICa~\cite{Yang2022PICa} utilize image captions as the bridge between visual model and GPT-3 to peform few-shot knowledge-based VQA. Socratic models~\cite{Zeng2022SocraticModels} present a modular framework that utilizes language-based exchange between pre-trained models and other modules. However, these studies only rely on text as the shared interface, which can inevitably lead to information loss when translating visual information into language. In addition, several concurrent studies \cite{wu2023visual,suris2023vipergpt,lu2023chameleon} have also explored the utilization of large language models for composing various language and visual models.

\paragraph{Question Decomposition}
Recent research has underscored the effectiveness of question decomposition and sub-question generation techniques in tackling complex tasks. DECOMPRC~\cite{QDmultiQA}, for instance, utilizes a limited amount of human-labeled data to train a span-based sub-question generator and simplifies multi-hop questions into single-hop questions. Similarly, \cite{RLQDRodrigo2017} leverages reinforcement learning for weakly supervised question generation and \cite{UnsupervisedQDPerez} introduces ONUS, an algorithm that harnesses large-scale questions sourced from the internet to perform unsupervised question decomposition. More recently, \cite{QD2022Pruthvi} proposes an alternative approach to enhance the performance of LLMs by decomposing challenging questions into simpler sub-questions on various tasks. Notably, the efficacy of question decomposition has been demonstrated across a range of tasks and domains, including solving mathematical problems~\cite{PDmath}, medical question answering~\cite{PDmedic}, and factual correction~\cite{QDfact}.

\section{Method}








\subsection{\method{}}


Figure~\ref{fig:overview} shows the overview of the \method{} approach, which is essentially a recursive thinking process involving a \textbf{\textit{top-down exploration process}} (in \textcolor{down_pink}{red} line) and a \textbf{\textit{bottom-up backtracking process}} (in \textcolor{back_green}{green} line). The top-down exploration process proactively breaks down the question into simpler sub-questions until the sub-questions are answered with high confidence. 
The bottom-up backtracking process recursively solves questions in which the answers to sub-questions are collected to solve the higher-level more complex questions.

In the beginning, we are given a target question $Q^{0,0}_1$, the context $C$ (if provided), and an optional hint $H^{0,0}_1$. The hint is initially Null but will be updated and enriched as the recursive thinking process continues and results from sub-questions are aggregated.
We first run the top-down process to explore the thinking space by invoking the {\selfQuestion} module. We use depth $d$ and turn $t$ to identify the node in our reasoning tree. Depth $d$ refers to the traditional depth of the recursion algorithm. Turn $t$ refers to the times of \method{} invoking the \selfQuestion{} module for each question.  For example, at depth $d$, turn $t$, 
\selfQuestion{} takes in the $i_{th}$ question $Q^{d,t}_i$, hint $H^{d,t}_i$, the context $C$, and decides if it can answer the question $Q^{d,t}_i$: (1) If \selfQuestion{} can directly output the answer $A^{d,t}_i$ for the question $Q^{d,t}_i$ with high confidence, the bottom-up backtracking process starts by converting the answer $A^{d,t}_i$ to a hint $\widetilde H^{d, t}_i$ with a QA-to-Hint module ($\widetilde H^{0, t}_i $ equals $ A^{0,t}_i$ directly when $d$ = 0) and adding $\widetilde H^{d, t}_i$
into the hints $H^{d-1,t}$ of the parent question $Q^{d-1}$. 
(2) If \selfQuestion{} cannot directly output an answer with high confidence, it outputs a set of sub-questions $\mathcal{Q}^{d+1,t}$ related to $Q^{d,t}_i$. Then we run \selfQuestion{} on each newly generated sub-question $Q^{d+1,t}_j$ until it's answered with high confidence. Once we obtain the answers to all the sub-questions $\mathcal{Q}^{d+1,t}$, we convert the answers into hints and incorporate them to update $H^{d,t}_i$ to $H^{d,t+1}_i$.
We then 
run \selfQuestion{} on $Q^{d,t+1}_i$ again with updated hints $H^{d,t+1}_i$. This recursive process continues until we reach the tree's root and the original question $Q^0_1$ is answered by $\tilde{H}^{0}_1$. We provide the pseudo-code of \method{} in Algorithm~\ref{alg:RMVP}.
\begin{algorithm}[!t]
\small
\caption{\method{} }\label{alg:RMVP}
\KwIn{Question $Q^{d,t}_i$, Hint $H^{d,t}_i$, Context $C$, Current Depth $d$, Max Depth $d_m$, Current Turn $t$, Max Turn $t_m$, Question Answer Prompt $P_{QA}$, Question Generate Prompt $P_{QG}$, QA to Hint Prompt $P_{\text{QA2H}}$}
\KwOut{Hint $\widetilde H^{d,t}_i$}

\For{$t \leq t_m$}{
    
    \tcp{call self-questioning}
    $<\mathcal{Q}^{d+1,t}, \mathcal{H}^{d+1,t}, C> \leftarrow \selfQuestion{}(Q^{d,t}_i, H^{d,t}_i, C,$
    $ d, d_m, t, t_m, P_{QA}, P_{QG})\;$;

    \uIf{$\mathcal{Q}^{d+1,t} \neq \emptyset$}{
        
        \For{each $Q^{d+1,t}_j \in \mathcal{Q}^{d+1,t}$}{
            \tcp{recursively answer sub-questions}
            
            $\widetilde H^{d+1,t}_j \leftarrow \text{\method{}}$($Q^{d+1}_j$, $H^{d+1,t}_j$, $C$, $d+1$, $d_m$, $t$, $t_m$, $P_{QA}$, $P_{QG}$)\;
            \tcp{gather hint}
            $H^d$.insert($\widetilde H^{d+1,t}_j$))\;
        }
    }\Else{
        $\widetilde H^{d,t}_j \leftarrow \mathcal{H}^{d+1,t}[0]$\;
        
        \textbf{return} $\widetilde H^{d,t}_j$\;
    }
    
    $t \leftarrow t+1$;
}
\end{algorithm}

\subsection{\selfQuestion{}}
\selfQuestion{} is designed to answer the given question, self-check the answer, and raise sub-questions. 
At depth $d$, turn $t$, \selfQuestion{} takes in the $i_{th}$ question $Q^{d,t}_i$, the context $C$ (if available), and hints $H^{d,t}_i$ (if available) and tries to generate an answer or a set of related sub-questions. \selfQuestion{} consists of two modules, a \textbf{Question-Answering (QA) Module} that outputs an answer $A^{d,t}_i$ 
for $Q^{d,t}_i$ based on $C$ and $H^{d,t}_i$, and an associated confidence level: high, medium, or low.
If the confidence of the answer is high, or either depth $d$ or turn $t$ met the pre-defined limit $d_m$ and $t_m$, \selfQuestion{} invokes the \QAtohint{} module to merge the question $Q^{d,t}_i$ and answer $A^{d,t}_i$ to hint $\widetilde H^{d,t}_i$ as output (when $d$ = 0, we skip the merging process because the answer $A^{0,}_1$ is the final answer and does not need to be rewritten to hint). 
Both Max Depth $d_m$ and Max Turn $t_m$ prevent \method{} from infinite recursion. 
On the other hand, if the confidence of the answer is lower than high, a \textbf{Question-Generation (QG) Module} is called to generate a set of sub-questions $\{Q_{0}^{d+1,t}, .., Q_{n}^{d+1,t}\}$ 
to collect more information based on $Q^{d,t}_i$, $C$, and $H^{d,t}_i$, where $n < n_m$ and $n_m$ denotes the maximum number of sub-questions to be generated. Algorithm~\ref{alg:self_question} shows the pseudo-code of the \selfQuestion{} algorithm. 

\begin{algorithm}[!t]
\small
\caption{\selfQuestion{}}\label{alg:self_question}
\KwIn{Question $Q^{d,t}_i$, Hint $H^{d,t}_i$, Context $C$, Current Depth $d$, Max Depth $d_m$, Current Turn $t$, Max Turn $t_m$, Question Answer Prompt $P_{QA}$, Question Generate Prompt $P_{QG}$, QA to Hint Prompt $P_{\text{QA2H}}$}
\KwOut{$<\mathcal{Q}^{d+1,t}, \mathcal{H}^{d+1,t}, C>$}
$\text{Must\_Answer} \leftarrow $\textbf{False};

\uIf{$d = d_m$ \textbf{or} $t = t_m$}{
    $\text{Must\_Answer} \leftarrow $\textbf{True};
}
\tcp{call the Question-Answering module}
$<A^{d,t}_i, confidence> \leftarrow QA{}(Q^{d,t}_i, H^{d,t}_i, C, P_{QA})\;$;

\uIf{$confidence = \text{high}$ \textbf{or} $\text{Must\_Answer}$}{
    \uIf{$d \neq 0$}{
        \tcp{merge QA to a hint}
        $\widetilde H^{d,t}_i \leftarrow \QAtohint{}(Q^{d,t}_i, A^{d,t}_i, P_{\text{QA2H}})\;$;
    }\Else{
        $\widetilde H^{d,t}_i \leftarrow A^d\;$;
    }
    $\mathcal{Q}^{d+1} \leftarrow \emptyset$;

    $\mathcal{H}^{d+1,t} \leftarrow \{\widetilde H^{d,t}_i\}$;

}\Else{
    \tcp{call the Question-Generation module}
    $\mathcal{Q}^{d+1,t} \leftarrow QG{}(Q^{d,t}_i, H^{d,t}_i, C, P_{QG})\;$;

    $\mathcal{H}^{d+1,t} \leftarrow \emptyset$;

}
\textbf{return} $<\mathcal{Q}^{d+1,t}, \mathcal{H}^{d+1,t}, C>$;

    
        

\end{algorithm}

\subsubsection{Question-Answering (QA) Module} \label{sec:QA}
\label{sec:qr_module}
The QA module aims to answer either the target question or a sub-question asked by the \selfQuestion{} module, based on the optional context and hints. We propose to leverage a large-scale language model (LLM), such as GPT-3 or ChatGPT~\cite{ChatGPT}, to answer the question given their superior reasoning capabilities demonstrated in previous studies~\cite{brown2020language, Zhang2022OPT, wei2022COT, Touvron2023LLAMA, yao2023ToT}.

Specifically, the input to the QA module consists of the given question $Q^{d,t}_i$, the context $C$, the optional hints $H^{d,t}_i$, and a prompt $P_{\text{QA}}$
designed to guide the QA module to generate an answer $A^{d,t}_i$ based on the inputs and output a confidence level. When the hints $H^{d,t}_i$ are available, $P_{\text{QA}}$ also asks the QA module to indicate which hints ared used to produce the answer.
\begin{equation} 
\label{eq:llm}
\small
    A^{d,t}_i, confidence =\text{QA}(Q^{d,t}_i, H^{d,t}_i, C, P_{QA}),
\end{equation}
where $confidence \in \{high, medium, low\}$. 

\subsubsection{Question-Generation (QG) Module} \label{sec:QG}
When the QA module outputs an answer for question $Q^{d,t}_i$ with low confidence, it's very likely that the answer is not correct and we need to collect additional hints to help the QA module produce a more confident answer. To do so, we design a Question-Generation (QG) module to raise a set of sub-questions that are related to $Q^{d,t}_i$. The QG module is also based on a large language model, such as ChatGPT, that takes the question $Q^{d,t}_i$, optional hints $H^{d,t}_i$, the context $C$, and a prompt $P_\text{QG}$ as input and outputs a set of sub-questions: 
\begin{equation}
\small
\begin{split}
    \{\mathcal{Q}_{0}^{d+1}, ..., \mathcal{Q}_{n}^{d+1}\}=\text{QG}(Q^{d,t}_i, H^{d,t}_i, &C, P_{QG}),
\end{split}
\end{equation}
where $n<n_m$. Intuitively, the sub-questions should be simpler than 
$Q^{d,t}_i$ and more likely to be answered by the QA module with high confidence.


\subsubsection{QA-to-Hint (QA2H) Module}
Since the answers to sub-questions may not be self-contained, we further design a QA-to-Hint module (QA2H) to merge each sub-question with its answer into a statement. Specifically, we feed the sub-question $Q^{d,t}_i$ and its answer $A^{d,t}_i$ to an LLM with the prompt $P_{\text{QA2H}}$ which asks the LLM to rewrite the question to a statement by incorporating the answer:
\begin{equation}
\small
    \tilde{H^{d}} = \QAtohint{}(Q^{d,t}_i, A^{d,t}_i, P_{\text{QA2H}}),
\end{equation}

\section{\method{} for Few-Shot Multimodal Reasoning}


\method{} can be naturally applied to text-based complex reasoning tasks as all the key components are based on large language models, such as ChatGPT. There are two critical challenges when applying \method{} to multimodal reasoning: (1) the language model cannot process visual information, and (2) simply applying a generic captioning model to convert visual content to natural language may not capture the key information required to answer a question.

\paragraph{Converting Visual Information into Context}
We propose to leverage LLMs to answer visual questions since some of the visual questions are knowledge-demanding~\cite{okvqa,schwenk2022okvqa} and LLMs are capable of storing commonsense knowledge and excel in complex reasoning tasks~\cite{brown2020language,wei2022COT, wang2022self}. 
To overcome the LLMs' shortcomings that they cannot perceive visual information, previous works~\cite{Yang2022PICa, Zeng2022SocraticModels} leverage an image captioning model to convert visual information into text and use LLMs to perform few-shot visual question answering (VQA) tasks. However, considering the richness and density of the information contained in an image, a generic caption may not be able to capture the key information that is necessary to answer a question. Thus, in order to adapt our \method{}, we employ a visual perception model, BLIP-2~\cite{blip2}, to describe the content of the image that is specific to a prompt. The input to BLIP-2 is an image $I$ (i.e., the image input of the VQA task) and a text prompt $Q$, and the output is an image caption $C$ describing the part of the image related to the prompt: $C =\text{BLIP-2}(I, Q)$, where
the text prompt $Q$ 
corresponds to $Q^d$ in Equation (\ref{eq:llm}) and the caption $C$ corresponds to the context $C$ in Equation (\ref{eq:llm}). 
By leveraging the visual perception model, we are able to resolve the hindrance and adopt our \method{} framework on VQA. We show more details on how we adapt \method{} to VQA in Appendix~\ref{app_vqa}.


\begin{table*}[ht!]
\centering
\resizebox{0.9\textwidth}{!}{%
\begin{tabular}{c|ccccc}
\toprule
\multicolumn{1}{l|}{}  & \textbf{MATH (DA)} & \textbf{MMLU Physics} & \textbf{MMLU Chemistry} & \textbf{LogiQA} & \textbf{Avg} \\\midrule
Standard-Prompting       & 7.00            & 65.11          & 53.20              & 54.67     & 45.00      \\
CoT~\cite{wei2022COT}           & 7.33            & 67.66          & 57.14              & 48.33     & 45.12       \\
SC-CoT~\cite{wei2022COT}           & 7.00            & 68.51          & 59.33              & 49.00    & 46.03        \\
ToT~\cite{yao2023ToT}           & 0.00            & 40.00       & 26.60              & 22.22     & 29.46       \\
\midrule
\method{} (2-Turns) & \underline{7.67}   & \textbf{71.49} & 63.55              & \textbf{59.33}  & \underline{50.51} \\
\method{} (3-Turns) & \textbf{11.67}            & \underline{69.36}          & \textbf{63.55}     & \underline{58.00}     & \textbf{50.65}      \\\bottomrule
\end{tabular}
}
\caption{Accuracy (\%) using Exact Match. The best performance is highlighted in \textbf{bold} and the second best performance is highlighted with \underline{underline}.} 
\vspace{-2mm}
\label{tab:result_text}
\end{table*}

\section{Experiment Setups}

\paragraph{Language-Only Tasks}
We leverage ChatGPT as the LLM for QA, QG, and QA2H modules, and provide detailed prompts for each module in Appendix \ref{app:prompt}. We evaluate \method{} on several complex reasoning tasks, including the Physics and Chemistry tasks in \textbf{Massive Multitask Language Understanding (MMLU)~\cite{Hendrycks2020multitask}}, Mathematical tasks in MATH~\cite{Hendrycks2021math}, and logical reasoning tasks based on LogiQA~\cite{Liu2020logiqa}. We adopt several state-of-the-art prompting methods as baselines, including \textbf{Standard Prompting (SP)} that directly prompts ChatGPT to answers a question with a few in-context examples. \textbf{Chain-of-Thought (CoT)~\cite{wei2022COT}}, \textbf{Self-Consistency Chain-of-Thought (SC-CoT)~\cite{wang2022self}}, and \textbf{Tree-of-Thought (ToT)~\cite{yao2023ToT}}. Following previous studies~\cite{chowdhery2022palm,hoffmann2022training}, we use exact match to measure the accuracy for all language-only tasks. More details for the baselines, evaluation metrics, and evaluation datasets are discussed in Appendix~\ref{appendix:lang}.

\paragraph{Multimodal Tasks}

We use blip2-flan-t5-xl as our \visualPercept{} module. We leverage ChatGPT~\cite{ChatGPT} for Factual/Visual Question Generation and Factual Question Answering and GPT-3 (GPT-3-davinci-003) for Visual Question Answering\footnote{These components are detailed in Appendix~\ref{app_vqa}.}, motivated by the observation that ChatGPT tends to be excessively cautious and neutral, and avoids answering some questions. We provide detailed sample prompts for each module in Appendix \ref{app:prompt}. We evaluate \method{} on several visual question answering datasets, including \textbf{VQA-V2~\cite{goyal2017making}}, \textbf{OK-VQA~\cite{okvqa}} and
\textbf{AOK-VQA~\cite{schwenk2022okvqa}}, and compare our approach with several baselines, including \textbf{\BLIP{}~\cite{blip2}} and.
\textbf{PICa~\cite{Yang2022PICa}}. More details for implementation, baselines, and datasets are discussed in Appendix~\ref{Appendix:multi}. For evaluation, we employ the conventional VQA accuracy metric \cite{goyal2017making} to measure the performance. To alleviate stringent penalization for minor discrepancies between predicted answers and ground truth, we normalize the answers by converting plural forms to singular forms and changing the tense of verbs to present tense. In addition, to address the conventional metric's limitation due to synonyms and expression differences, we design semantic-based accuracy by employing ChatGPT to evaluate the correctness of the predicted answers~\cite{Fu2023GPTscore, Liu2023Geval}. We provide ChatGPT with the visual question, the predicted answer and the ground-truth answer, and ask if the ground-truth answer and the predicted answer can support each other. If the answer is "Yes", we treat the predicted answer as correct. We show the exact prompts used for ChatGPT in Appendix \ref{app:grade}.

\section{Results and Discussions}

\subsection{Quantitative Results}
\label{sec_quant}
\begin{figure*}[ht!]
    \centering
    \includegraphics[width=0.9\textwidth]{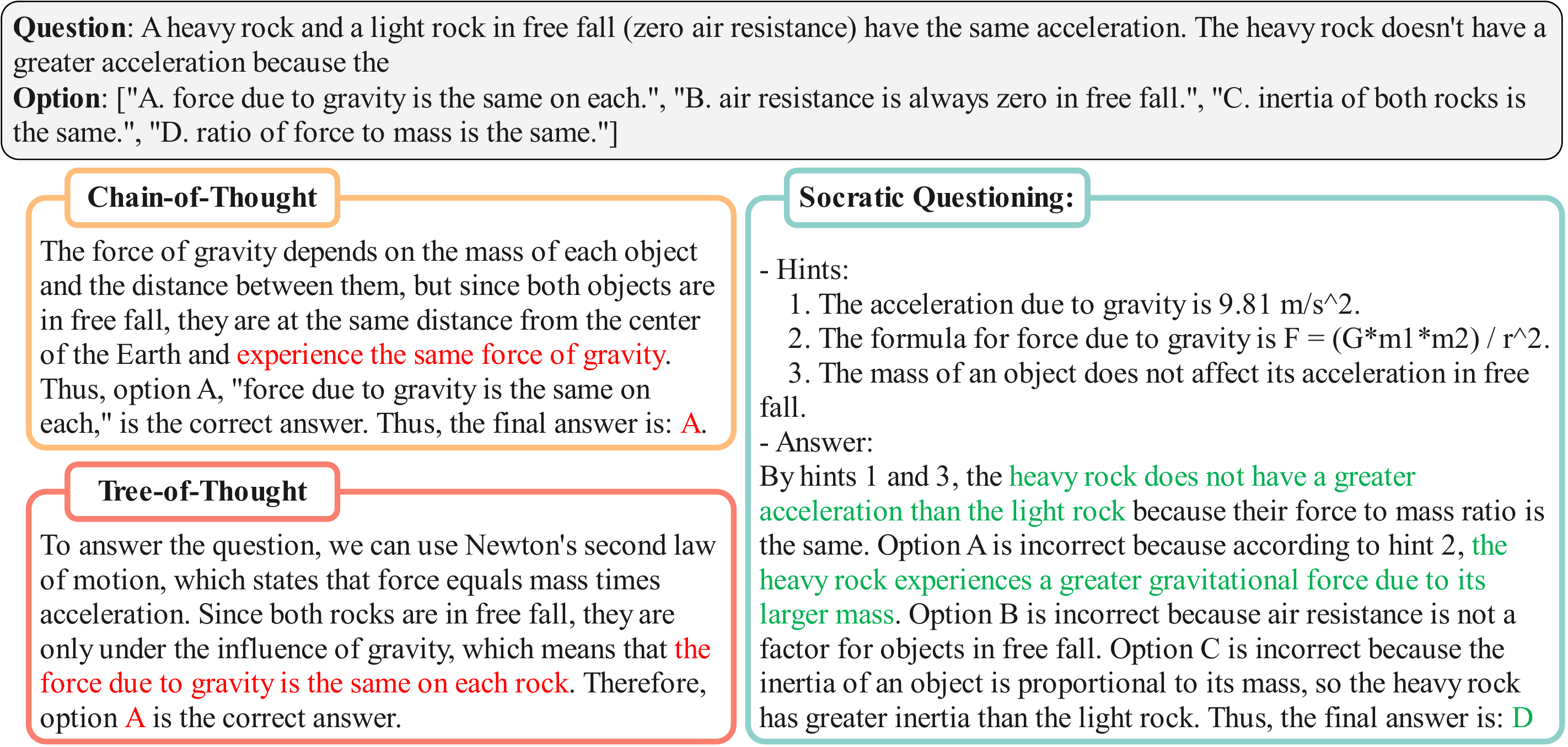}
    \caption{Qualitative results of CoT, ToT, and \method{} on the Physics task. The correct answer of this example is \textbf{D}.}
    \label{fig:textresult_exp}
\end{figure*}
\paragraph{Language-only Tasks} 
Table~\ref{tab:result_text} shows the quantitative results in terms of accuracy for language-only reasoning tasks. Our method substantially outperforms previous state-of-the-art methods by 4.34\%, 2.98\%, 4.22\%, and 4.66\% absolute gains in MATH, Physics, Chemistry, and Logic benchmarks, respectively. This effectively demonstrates the superiority of our approach. 
We also conduct an experiment on how the maximum number of turns $t_m$ affects the performance. Specifically, we experiment with the setting where $t_m=2$ (2-Turns) and $t_m=3$ (3-Turns). From Table~\ref{tab:result_text}, the model with maximum 2 turns achieves better performance on Physics and LogiQA datasets, while the model with $t_m=3$ performs better on the MATH dataset. 
One possible reason is that the Physics and LogiQA benchmarks may not be challenging enough and reasoning within 2 turns is sufficient to answer most of the questions. We provide a concrete example in Appendix~\ref{apd:concrete_turn}.



\begin{table}[t!]
\centering
\resizebox{0.5\textwidth}{!}{%
\begin{tabular}{c|ccc}
\toprule
\textbf{Model} & \textbf{VQA-V2} & \textbf{OK-VQA} & \textbf{AOK-VQA} \\ \midrule
\BLIP{}~\cite{blip2}                     & 36.7            & 21.14           & 0                \\
PICa~\cite{Yang2022PICa}                     & \underline{43.18}     & \underline{29.94}     & \underline{28.6}       \\
\method{}       & \textbf{46.64}  & \textbf{31.24}  & \textbf{29.58}   \\ \bottomrule
\end{tabular}
}
\caption{Traditional VQA Accuracy (\%) based on Exactly Match. The best performance is highlighted in \textbf{bold} and the second best performance is highlighted with \underline{underline}. } 
\label{tab:result_em}
\end{table}

\paragraph{Multimodal Tasks} Table \ref{tab:result_em} and \ref{tab:result_sm} show the quantitative results using traditional VQA accuracy and semantic-based accuracy, respectively. 
For both results, our \method{} method outperforms the previous state-of-the-art approaches on most benchmarks, often by a large margin. The only exception is semantic-based accuracy on the VQA-V2 dataset. A possible reason is that the tasks on VQA-V2 focus more on the visual recognition and detection aspect and do not require much reasoning capability and external knowledge. 

\begin{table}[t!]
\centering
\resizebox{0.5\textwidth}{!}{%
\begin{tabular}{c|ccc}
\toprule
\textbf{Model}  & \textbf{VQA-V2} & \textbf{OK-VQA} & \textbf{AOK-VQA} \\ \midrule
\BLIP{}~\cite{blip2}                     & \textbf{57.2}            & 46.75           & 43.29                \\
PICa~\cite{Yang2022PICa}                     & 49.8     & \underline{48.05}     & \underline{46.85}       \\
\method{}                    & \underline{54.4}  & \textbf{53.03}  & \textbf{49.55}   \\ \bottomrule
\end{tabular}
}
\caption{Semantic-based VQA Accuracy (\%) using NLI. The best performance is highlighted in \textbf{bold} and the second best performance is highlighted with \underline{underline}.} 
\vspace{-4mm}
\label{tab:result_sm}
\end{table}

\begin{figure}[ht!]
    \centering
    \includegraphics[width=0.48\textwidth]{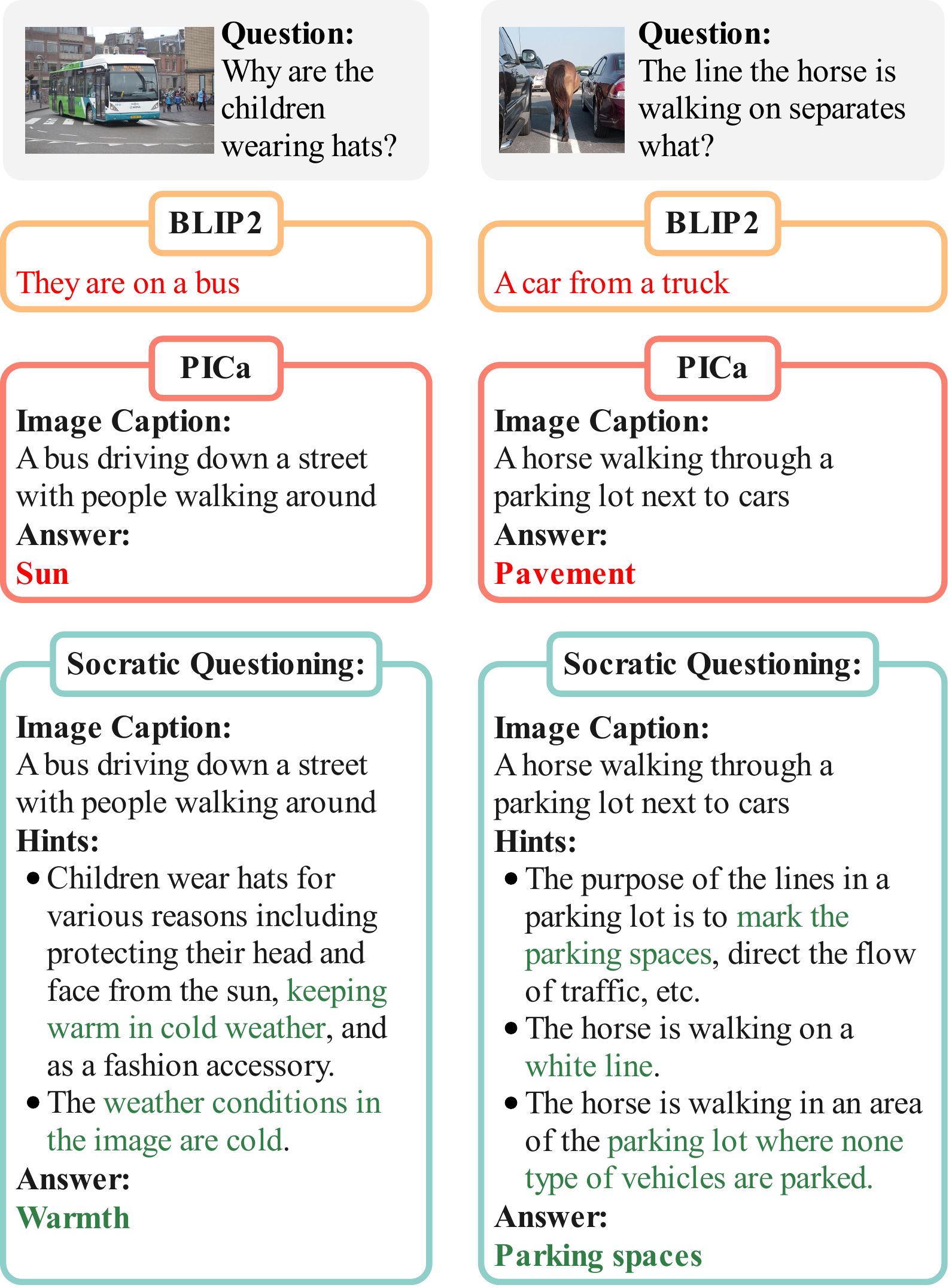}
    \caption{Qualitative results of few-shot VQA using \BLIP{}, PICa, and \method{} (2-Depth 2-Turn).}
    \label{fig:result_exp}
    \vspace{-3mm}
\end{figure}

\begin{figure}[h]
    \centering
    \includegraphics[width=0.43\textwidth]{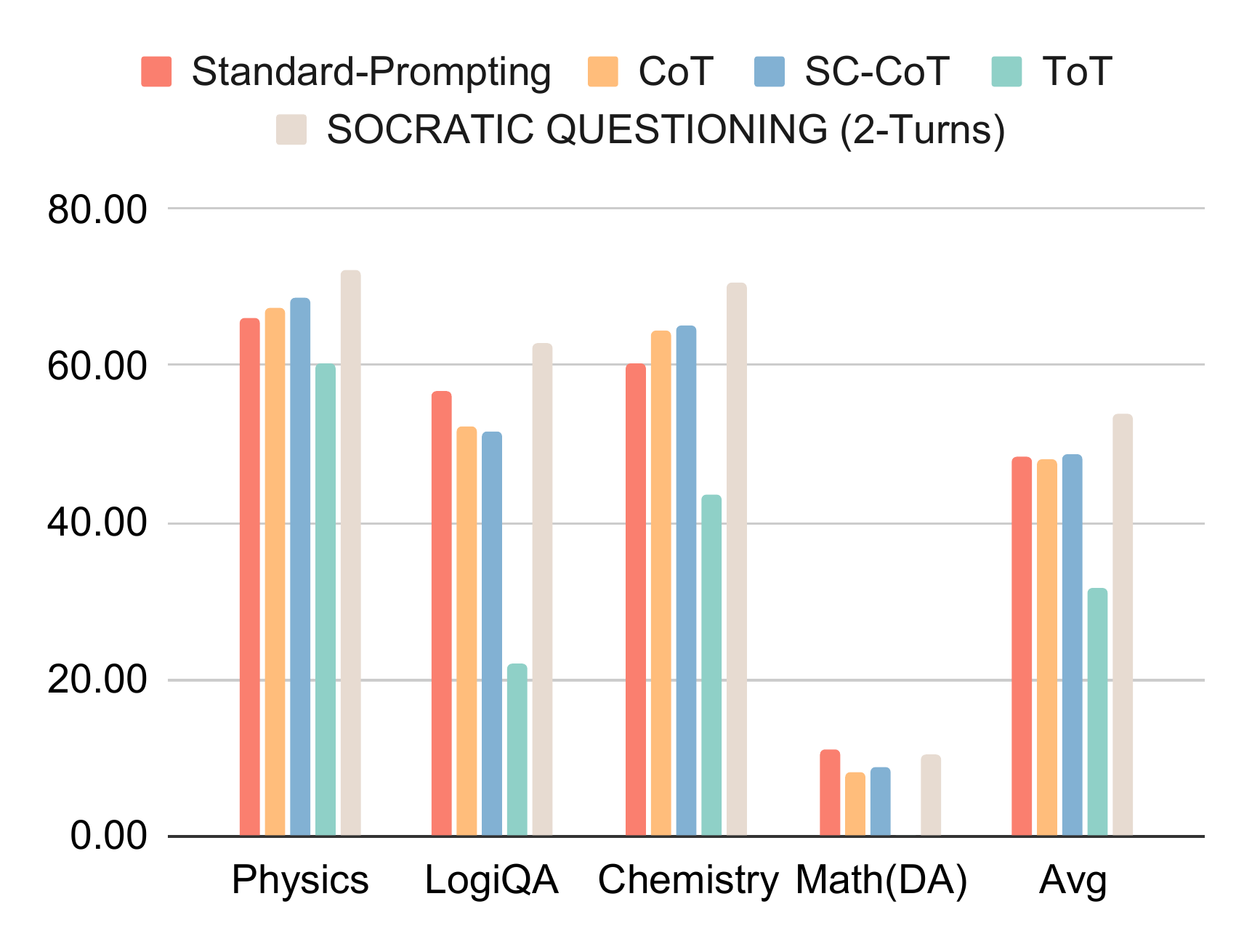}
    \caption{Accuracy (\%) on the examples that triggered 2 turns of reasoning by \method{}.}
    \vspace{-3mm}
    \label{fig:result_2turn}
\end{figure}

\begin{figure}[h]
    \centering
    \includegraphics[width=0.43\textwidth]{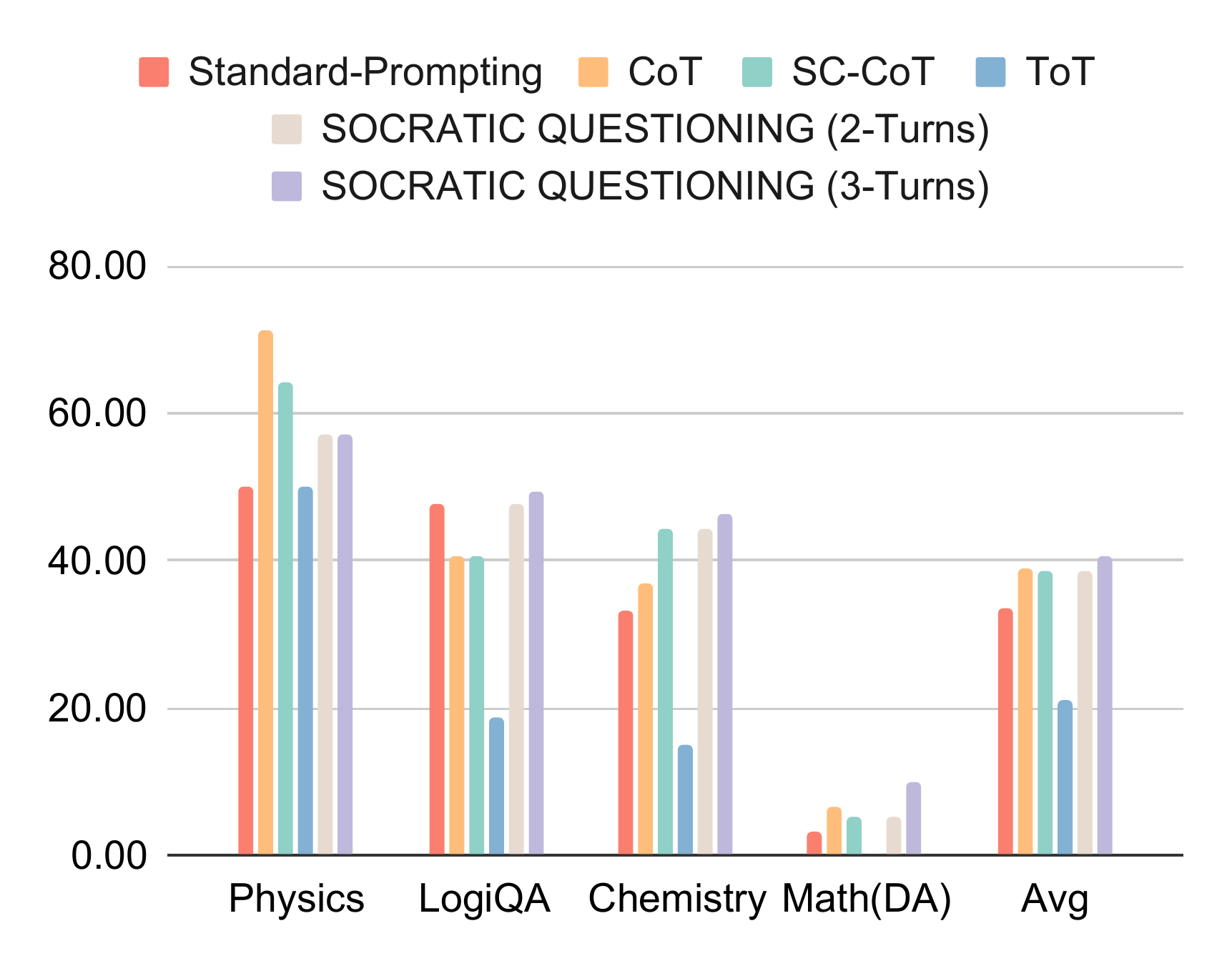}
    \caption{Accuracy (\%) on the examples that triggered 3 turns of reasoning by \method{}.}
    \vspace{-4mm}
    \label{fig:result_3turn}
\end{figure}

\begin{figure}[h]
    \centering
    \includegraphics[width=0.27\textwidth]{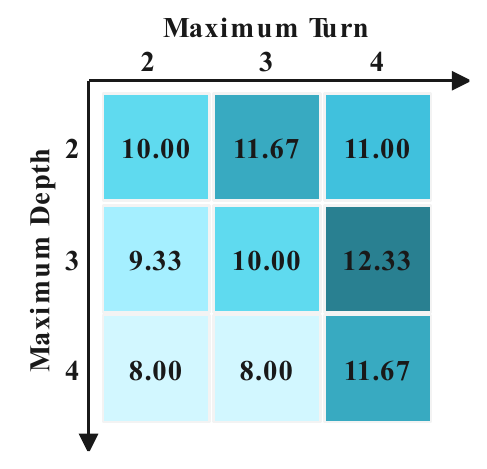}
    \caption{Quantitative results \method{} on the MATH dataset with different values of the hyperparameters $t_m$ and $d_m$.}
    \vspace{-3mm}
    \label{fig:result_heatmap}
\end{figure}







\subsection{Qualitative Result}
\paragraph{Language-only Tasks} 
Figure~\ref{fig:textresult_exp} shows the qualitative results of \method{} and baselines on the Physics task. As one can observe, \method{} can effectively prompt hints containing the necessary information to solve the original problem and selectively use the hints to reach the correct final answer. On the other hand, CoT and ToT reach the wrong answer due to the poorly sampled reasoning path.

\paragraph{Multimodal Tasks} 
Figure~\ref{fig:result_exp} shows several examples of few-shot VQA tasks from the baselines and \method{}. We demonstrate that the hints acquired via the sub-problems are highly related to the original problem (e.g., \textit{"weather conditions are cold"}), and by considering the collected hints, the \method{} reaches the correct final answer (e.g., \textit{"warmth"}). In contrast, the answer from \BLIP{} is irrelevant to the given question, due to the generic caption.

\begin{table}[t!]
\centering
\resizebox{0.49\textwidth}{!}{%
\begin{tabular}{c|c|c}
\toprule
              & Answered Correctly & Answered Incorrectly  \\ \midrule
Avg. Hints & 3.28           & 3.68             \\
Avg. Depth & 2.89          & 2.92             \\ \bottomrule
\end{tabular}
}
\caption{Averaged numbers of hints and depth of \method{} used for questions answered correctly and incorrectly, respectively. } 
\label{tab:hint_distribute}
\end{table}

\subsection{How do the Numbers of Turns and Depths Affect the Model?}

\paragraph{Performance Breakdown w/ Number of Turns} To study how the number of reasoning turns affects the performance across different benchmarks, we investigate how the baselines and our method perform on the examples that triggered 2 and 3 turns of reasoning by \method{} in Figure~\ref{fig:result_2turn} and  Figure~\ref{fig:result_3turn}, respectively. This experiment can be considered as breaking down the results in Table~\ref{tab:result_text} into two groups based on the number of reasoning turns. From Figure~\ref{fig:result_2turn}, our approach outperforms the baselines on all benchmarks except for the MATH dataset. From Figure~\ref{fig:result_3turn}, our approach outperforms the baselines on relatively challenging tasks such as MATH but performs more poorly on easier tasks such as Physics. This indicates \method{} with more turns can tackle challenging problems more effectively.

\paragraph{The Effect of Hyperparameters $t_m$ and $d_m$}
In addition to the discussion in~\ref{sec_quant}, we conduct a more in-depth analysis of how the maximum number of turns $t_m$ and maximum number of depths $d_m$ affect the performance of our \method{}. In Figure~\ref{fig:result_heatmap}, we show the heat map under different hyperparameter settings, where the number in each cell is the accuracy (\%) given a specific combination of $t_m$ and $d_m$. We observe two general trends: (1) the accuracy increases when $t_m$ gets larger, and (2) the accuracy decreases when $d_m$ gets larger. These results imply that our approach can benefit from raising more questions directly related to the original question. Also, performing reasoning with a larger maximum depth does not yield better performance since the benchmark may not be challenging enough, and exploring at a deeper level may introduce irrelevant information. We provide a concrete example in Appendix~\ref{apd:concrete_depth}. In addition, we analyze the computational cost of \method{} compared to other baselines in Appendix~\ref{apd:cost}, and show that while achieving stronger performance, our proposed algorithm enjoys higher efficiency than most of baselines.

\subsection{How does the Difficulty of Questions Affect the Model?}
Table~\ref{tab:hint_distribute} presents the averaged numbers of hints and depth used to answer the original questions for correct and incorrect answers. As one can observe, for incorrect answers, the LLM raises more sub-questions, which demonstrates that the LLM tends to explore more thinking space when tackling questions that it does not know the answers. This trend also agrees with the depth. If the question is hard for the LLM, the model tends to break the sub-questions into even more basic questions.

\section{Conclusion}
We present \method{}, a novel divide-and-conquer fashion algorithm that is inspired by human's recursive thinking processes. 
\method{} consists of a top-down reasoning phase that decomposes a complex problem into simpler sub-problems and a bottom-top phase where the solutions to the sub-problems are recursively returned and used to solve the original problem at higher levels. 
Extensive experiments on four challenging language-only tasks and the few-shot VQA task validate the effectiveness of our \method{}. Moreover, qualitative analysis demonstrates our approach can effectively elicit intermediate reasoning steps and consequently yield a correct final answer while enjoying transparency and interpretability. 
\section*{Limitation}
The self-checking functionality lacks sufficient sensitivity to incorrect responses, as its confidence estimation heavily relies on LLMs themselves. While we employed ChatGPT as the backbone for our algorithm, its tendency towards overconfidence leads to a low frequency of sub-question generation. 

Our study exhibits a lack of diversity in visual models used to extract information from images. We only use BLIP-2~\cite{blip2} as an image caption model in current experiments. However, the incorporation of diverse visual models, such as dense caption models, Optical Character Recognition (OCR), or scene graph models, may potentially yield a broader spectrum of image information, thus facilitating the resolution of sub-questions. In addition, to help BLIP-2 to better follow instructions from LLMs, we propose to leverage recent techniques developed in visual instruction tuning~\cite{liu2023llava,multiinstruct,visionFlan2023,instructblip}.

Additionally, our experiments were constrained to the English language datasets and we only consider the VQA task to showcase the multi-modal performance. However, given the generality of our algorithm, we plan to test its functionality with multilingual datasets and experiment it on other domains, such as speech~\cite{speechdata, you2022end}, and video~\cite{visual-CoT}.
\section*{Acknowledgments}
This research is based upon work supported by the U.S. DARPA ECOLE Program \# HR001122S0052. The views and conclusions contained herein are those of the authors and should not be interpreted as necessarily representing the official policies, either expressed or implied, of DARPA or the U.S. Government. The U.S. Government is authorized to reproduce and distribute reprints for governmental purposes notwithstanding any copyright annotation therein.

\bibliography{anthology,custom}
\bibliographystyle{acl_natbib}

\appendix

\section{Adapting \method{} to Visual Question Answering}
\label{app_vqa}
\paragraph{Question-Generation (QG) Module}
Some tasks (e.g., OK-VQA, AOK-VQA) require commonsense knowledge. Although LLMs can retrieve knowledge from its parameter, they are prone to hallucination and the black-box retrieving process is hard to debug. In order to gain a clear understanding of the factual knowledge used in answering a question, we divide the QG module in Section~\ref{sec:QG} into two sub-modules: A Fact-Question-Generation (FQG) sub-module which generates factual questions related to background knowledge of the given question, and a Visual-Question-Generation (VQG) sub-module generates visual questions, which aims to guide the Visual Perception module to focus on question-related image regions and seek more image information.

\paragraph{Question-Answering (QA) Module}
To accommodate the two question types, we also divide the QA module in section~\ref{sec:QA} into two sub-modules: A Factual-Question-Answering module (FQA) and a Visual-Question-Answering module (\VQA{}). Both FQA and \VQA{} modules follow the same formulation in Equation (\ref{eq:llm}). 
The input $C$ to \VQA{} is the caption related to the question $Q^d$ and is prompted via the Equation of BLIP-2.
\paragraph{\selfQuestion{}}
Figure~\ref{fig:selfq} demonstrates the detailed step of the \selfQuestion{} algorithm in the multimodal setting. At depth $d$, \selfQuestion{} algorithm takes in a visual question $Q^d$ which can be the original visual question ($d=0$) or a sub-question generated by \VQG{}, a question-related caption $C$, and hints $H^d$ (if it is available), and try to generate an answer $A^d$ via \VQA{}. If the confidence level of $A^d$ is not high, the
\selfQuestion{} algorithm starts to raise sub-questions. First, the \FQG{} module takes in $Q^d$, context $C$, and hints $H^d$ as input and raises a set of factual questions $\mathcal{Q}_f$. Each question in $\mathcal{Q}_f$ is answered by the \FQA{} module and we denote the answer as $A_f$. Each $Q_f$ and its answer $A_f$ is mearged into a factual statement $h_f$ via the QA2H module and the statement is appended to hints $H^d$ to form $H^{d+1}$. Second, the \VQG{} module takes in $Q^d$, context $C$, and hints $H^{d+1}$ and raises a set of visual questions $\mathcal{Q}^{d+1}$.

\section{Visualization of Recursive Thinking Process}
\label{app:thinking}
Figure \ref{fig:result_detail} shows a complete recursive thinking process of our \method{} method. It involves 4 additional questions to acquire additional information to answer the target question. From this example, we see that LLMs, such as GPT-3 or ChatGPT, have strong capabilities not only in reasoning but also self-questioning. 
Given the target question to be answered, ``\textit{Why are the children wearing hats?}'',
LLMs are able to proactively acquire additional commonsense knowledge through factual questions, e.g., ``\textit{What are the common reasons why children wear hats?}'', and fine-grained visual information from the input image, e.g., ``\textit{What's the position of the sun in the sky at the time the children are shown wearing hats}'', ``\textit{Are the weather conditions in the image cold or hot}''. By combining the additional knowledge, e.g., ``\textit{cold weather makes people wear hats}'' and visual information, e.g., ``\textit{it is cold}'', acquired from the recursive Self-Questioning process, the model finally achieves the answer ``\textit{warmth}''. 
This analysis demonstrates that the recursive thinking process of our approach is highly transparent and interpretable.


\section{Implementation Details}
\subsection{Language-only Tasks}
\label{appendix:lang}
\paragraph{Implementation Details}
We leverage ChatGPT~\cite{ChatGPT} as the LLM for QA, QG, and QA2H modules. We provide detailed prompts for each module in Appendix \ref{app:prompt}.
\paragraph{Baselines}
\textbf{Standard Prompting (SP)} prompts ChatGPT to directly answers a question with a few in-context examples. \textbf{Chain-of-Thought (CoT)~\cite{wei2022COT}} prompts ChatGPT to first generate the thinking process and then generate the answer. We also add the thinking process into the in-context examples.
\textbf{Self-Consistency Chain-of-Thought (SC-CoT)~\cite{wang2022self}} proposes to run chain-of-thought multiple times on ChatGPT and marginalize the thinking process by taking the most consistent answer. \textbf{Tree-of-Thought (ToT)~\cite{yao2023ToT}} is a recently proposed framework for improving the reasoning capability of language models. We follow their implementation~\footnote{\url{https://github.com/kyegomez/tree-of-thoughts}} which leverages tree-search algorithms to explore the thinking space and select the best thinking path.~\footnote{By the time we submit the work, we don't have access to GPT4 so we use ChatGPT for ToT.}

\paragraph{Evaluation Metrics}
For a fair comparison, we use exact match and measure the accuracy for all language-only tasks following previous works~\cite{chowdhery2022palm,hoffmann2022training}.

All questions in MMLU Physics, MMLU Chemistry, and LogiQA are multiple-choice questions and the answer is always a single letter like “A”, “B” or “C”. To easily parse the model’s final output, we use “Thus, the final answer is:” as the prefix for the final answers (A or B or C or D, ect.) in the in-context examples for all methods. When we parse the output, we first run a template-based method to extract the answers after “Thus, the final answer is:”. For a few instances (12.52\% in CoT, 16.4\% in ToT and 11.64\% in Socratic Questioning on average) that do not match the template as shown in Figure~\ref{fig:textresult_exp} ToT, the authors manually compare the model’s predictions to the ground truth answers. Thus, we assure that the final performance of all methods is not affected by the output formats.

\paragraph{Datasets}
\textbf{Massive Multitask Language Understanding (MMLU)~\cite{Hendrycks2020multitask}} dataset contains  57 diverse tasks and is used to measure the model's complex reasoning capability. In this work, we use the physics and chemistry tasks which contain conceptual physics and chemistry multiple-choice questions, respectively.
\textbf{MATH~\cite{Hendrycks2021math}} dataset consists of challenging competition-level mathematics problems which require strong mathematical reasoning ability.
\textbf{LogiQA~\cite{Liu2020logiqa}} dataset contains expert-written questions for testing the logical reasoning capability of humans.
For each task, we use the validation set to make design decisions and measure the model's performance on the test set. The detailed statistics of all datasets can be found in Table~\ref{tab:textdata_statistic}.


\begin{table}[]
\centering
\resizebox{0.5\textwidth}{!}{%
\begin{tabular}{c|cccc}
\toprule 
\multicolumn{1}{l|}{} & \textbf{MATH} & \textbf{\begin{tabular}[c]{@{}c@{}}MMLU\\ (Physics)\end{tabular}} & \textbf{\begin{tabular}[c]{@{}c@{}}MMLU\\ (Chemistry)\end{tabular}} & \textbf{LogiQA} \\ \midrule
\textbf{Dev}          & 60            & 22                                                                & 26                                                                  & 60              \\
\textbf{Test}         & 300           & 235                                                               & 203                                                                 & 300             \\ \bottomrule 
\end{tabular}
}
\caption{Statistic of datasets for language-only tasks.} 
\label{tab:textdata_statistic}
\end{table}

\subsection{Multimodal Tasks}
\label{Appendix:multi}
\paragraph{Implementation Details}
We use blip2-flan-t5-xl\footnote{\url{https://huggingface.co/Salesforce/blip2-flan-t5-xl}} as our \visualPercept{} module.
We leverage ChatGPT~\cite{ChatGPT} for the FQG, VQG, and FQA modules and GPT-3 (GPT-3-davinci-003) for the VQA module. This decision is motivated by the observation that ChatGPT tends to be excessively cautious and neutral, and avoids answering some questions. We provide detailed sample prompts for each module in Appendix \ref{app:prompt}.

\paragraph{Baselines}
\textbf{\BLIP{}~\cite{blip2}}  
is a pre-trained vision-language model that leverages an efficient and generic pre-training strategy and is able to follow text prompts. We use the released blip2-flan-t5-xl checkpoint.
\textbf{PICa~\cite{Yang2022PICa}} prompts GPT-3 with generic image captions to solve VQA in an in-context learning manner. 
In our experiments, we implement PICa by using  blip2-flan-t5-xl as the image captioning model and \texttt{GPT-3-davinci-003} as the LLM.

\paragraph{Evaluation Metrics}
We employ the conventional VQA accuracy metric \cite{goyal2017making} to measure the performance. To alleviate stringent penalization for minor discrepancies between predicted answers and ground truth, we normalize the answers by converting plural forms to singular forms and changing the tense of verbs to present tense.
In addition, to address the limitation due to synonyms and expression differences, we employ ChatGPT to evaluate the correctness of the predicted answers~\cite{Fu2023GPTscore, Liu2023Geval}. We provide ChatGPT with the visual question, the predicted answer and the ground-truth answer, and ask if the ground-truth answer and the predicted answer can support each other. If the answer is "Yes", we treat the predicted answer as correct.
We show the exact prompts used for ChatGPT in Appendix \ref{app:grade}.

\paragraph{Datasets}
\textbf{VQA-V2~\cite{goyal2017making}}
is a dataset containing open-ended questions about images. 
\textbf{OK-VQA~\cite{okvqa}} requires model to leverage external knowledge to answer visual questions.
\textbf{AOK-VQA~\cite{schwenk2022okvqa}}
is an augmented successor of OK-VQA, which require commonsense knowledge and strong reasoning capabilities to answer its questions.
For each task, we use the validation set to make design decisions and measure the model's performance on the test set. The detailed statistics of all datasets can be found in Table~\ref{tab:visualdata_statistic} and Appendix \ref{apx:data_leak}.

\begin{table}[]
\centering
\resizebox{0.4\textwidth}{!}{%
\begin{tabular}{c|ccc}
\toprule 
\multicolumn{1}{l|}{} & \textbf{VQA-v2} & \textbf{OK-VQA} & \textbf{AOK-VQA} \\\midrule
\textbf{Dev}          & 100             & 100             & 100              \\
\textbf{Test}         & 500             & 462             & 444                  \\ \bottomrule 
\end{tabular}
}
\caption{Statistic of datasets for multi-modalities tasks.} 
\label{tab:visualdata_statistic}
\end{table}

\section{\selfQuestion{} in the Multimodal Setting}
\begin{figure*}[h]
    \centering
    \includegraphics[width=0.98\textwidth]{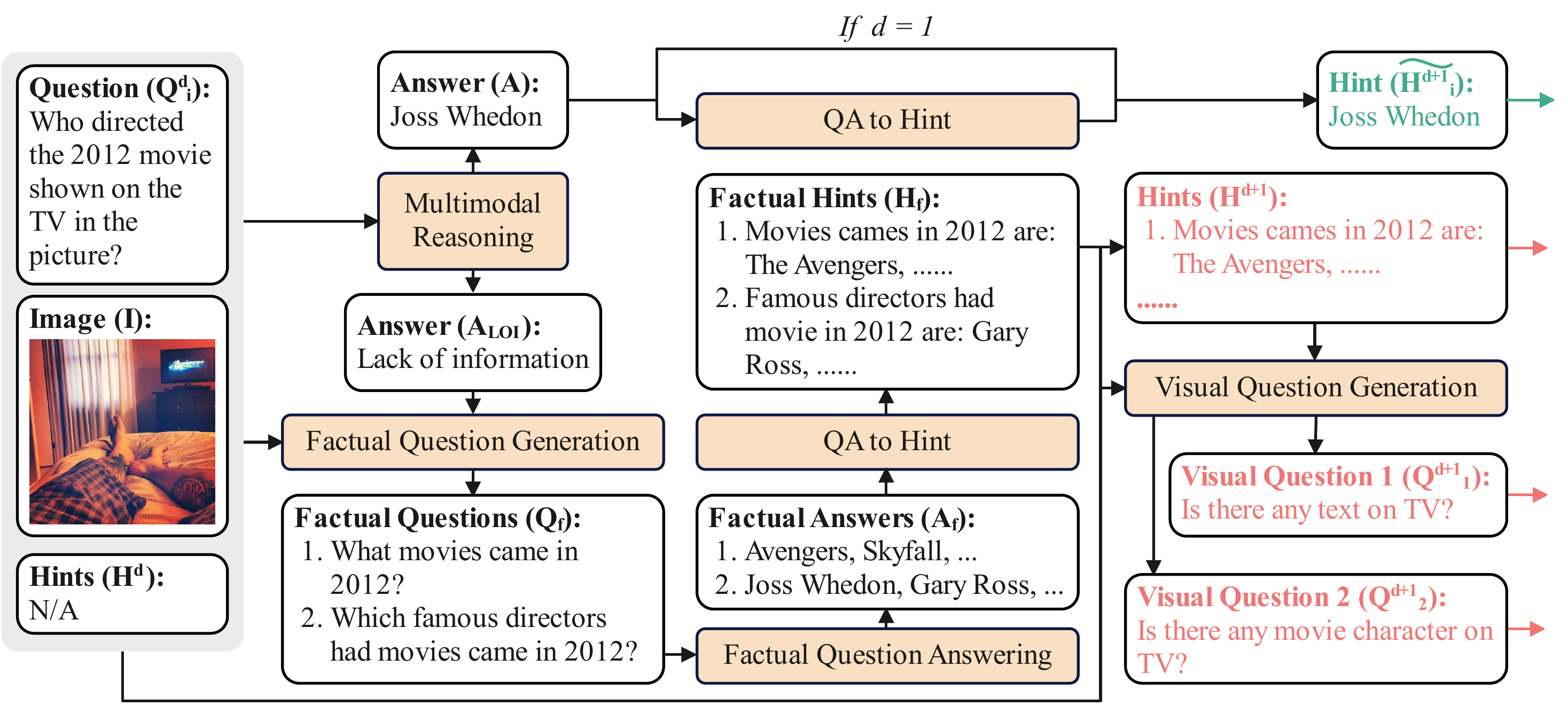}
    \caption{The overview of the \selfQuestion{} Algorithm.}
    \label{fig:selfq}
\end{figure*}

See Figure \ref{fig:selfq}.

\section{Data Leakage in BLIP-2 and GPT-3 }
\label{apx:data_leak}
\begin{table}[h]
\centering
\resizebox{0.45\textwidth}{!}{%
\begin{tabular}{c|c|c|c}
\toprule 
\textbf{Model} & \textbf{VQA-V2} & \textbf{OK-VQA} & \textbf{AOK-VQA} \\ \midrule
\BLIP{}          & 1.46            & 2.93            & 28.08            \\
GPT-3   & 35.88           & 23.95           & 20.4            \\ \bottomrule 
\end{tabular}
}
\caption{Traditional VQA Accuracy (\%) under the setting where no image is provided in the input.} 
\label{tab:no_img}
\end{table}
In our preliminary experiments, we discovered an issue that pre-trained models could be subject to data leakage during their pre-training stage. We observed that the baseline models (i.e., \BLIP{} and GPT-3) achieved unjustifiably high performance across all three VQA datasets even without taking images as inputs (see Table~\ref{tab:no_img}). 
To address this issue, we applied a filtering process to remove such contaminated instances. We first test the \BLIP{} and GPT-3 on zero-shot VQA tasks while replacing the original input image with an image composed entirely of black pixels of the same size. Then, we only retain the samples where the models failed to yield a correct answer when the original image is not given. After the filtering, we adopt the 500, 462, and 444  test samples for VQA-V2, OK-VQA, and AOK-VQA, respectively. We use these clean examples for the evaluation throughout the rest of our experiments. 

\section{Visualization of Complete \method{}}
\begin{figure*}[th]
    \includegraphics[width=0.98\textwidth]{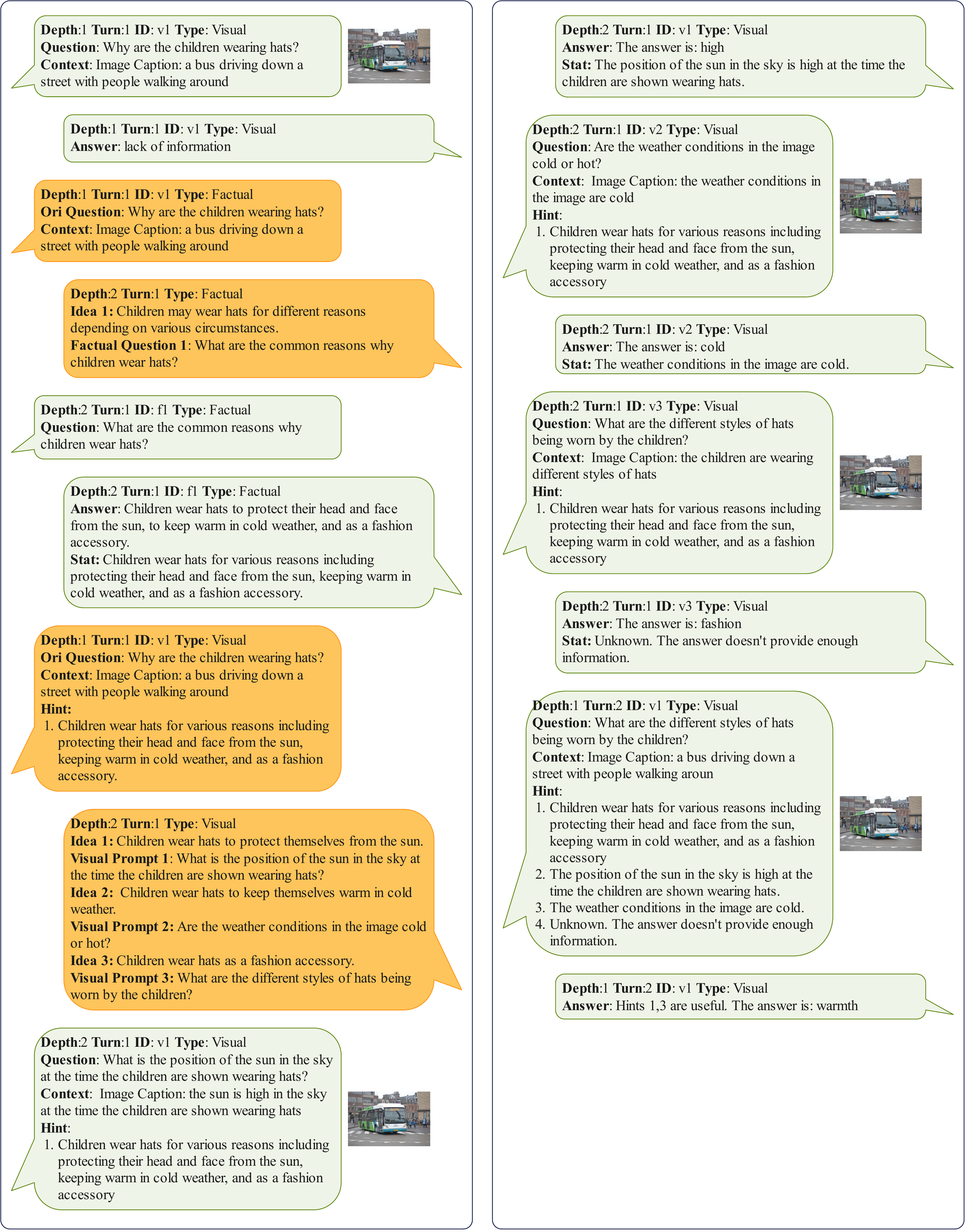}
    \caption{Visualization of a complete recursive thinking process of \method{} (2-Depth 2-Turn). The process is initialized on the left and is continued on the right.}
    \label{fig:result_detail}
\end{figure*}

See Figure \ref{fig:result_detail}.

\section{Concrete Example}
\subsection{Large Maximum Number of Turn}
\label{apd:concrete_turn}
Due to the calibration error in LLMs~\cite{jiang-etal-2021-know}, sometimes the pre-trained model’s confidence is not aligned with the answer’s correctness. Thus, in such cases, the model predicts “low” or “medium” confidence in correct answers in the early turns and hence misses the correct answers. If we use fewer turns, we can keep the answer in the early turn regardless of the confidence and hence alleviate the calibration error. Below we show a concrete example in which the model predicts the correct answer in 2 turns and predicts the incorrect answer in 3 turns. When we increase the number of turns, Socratic Questioning may raise some less relevant sub-questions and hence introduce noisy information in the reasoning process. This noisy information can confuse the model, leading to incorrect responses to the original question. For example, consider a simple physics question:

\textit{The speed of sound is slightly greater on a [ "A. cold day", "B. hot day", "C. day with steady temperature", "D. None of these"]?}

In a 2-turn setting, our approach obtains hints: (1) "The speed of sound increases with increasing temperature.", and (2) "Humidity is a factor in the speed of sound." According to the hints, it is obvious that the correct answer is B, which is chosen by our approach in the second turn with the "middle" confidence.
In a 3-turn setting, since the LLM does not assign “high” confidence to the answer in the 2 turn, our approach goes deeper in the third turn and gets more information (e.g., (3) "The speed of sound can be affected by several factors, including temperature, humidity and density of the medium.", (4) "The speed of sound depends on the density and elasticity of the medium it is traveling through, in terms of physical properties.", (5) "The speed of sound increases with humidity as a result of increased air density.") As a result, by considering more hints, we potentially introduce less relevant information to the LLM and the noisy information causes the LLM to change its answer to D.

\subsection{Large Maximum Number of Depth}
\label{apd:concrete_depth}
We observe that as the depth increases, the context information in the original questions start to vanish and the answers to the sub-questions may be inaccurate in the context of the original question. Thus, by adding the answers to sub-question in larger depth as hints, we can introduce noises to the reasoning process of the LLM which results in wrong answers. Consider a physics question example:

\textit{ When a spinning system contracts in the absence of an external torque, its rotational speed increases, and its angular momentum [ A. decreases, B. increases, C. remains unchanged, D. may increase or decrease ]"? }

Socratic Questioning raises a sub-question: \textit{"What affects the rotational speed of a spinning system?"} The initial answer to this sub-question is \textit{“Conservation of angular momentum”}, which provides enough information to answer the original question. In a larger depth setting, the Socratic Questioning raises a deeper sub-question: \textit{“What is the relationship between rotational speed and angular momentum in a spinning system?”} The answer to this question is: \textit{“The angular momentum is directly proportional to the rotational speed”}. Incorporate this hint, the Socratic Questioning changes the answer of the first sub-question to: \textit{“The angular momentum is directly proportional to the rotational speed.”}, which results in an incorrect final answer B.

\section{Evaluation of Computational Cost}
\label{apd:cost}

\begin{table*}[ht!]
\centering
\resizebox{\textwidth}{!}{%
\begin{tabular}{p{4.4cm}p{1.5cm}p{1cm}p{1cm}p{2.2cm}p{3.5cm}p{3.5cm}}
\toprule
                                           & Standard-Prompting & CoT  & SC-CoT & ToT           & \begin{tabular}[c]{@{}c@{}}Socratic Questioning \\ (2 turns)\end{tabular} & \begin{tabular}[c]{@{}c@{}}Socratic Questioning \\ (3 turns)\end{tabular} \\ \midrule
Theoretical Number of Calls                & 1                  & 1    & 20     & k + b*k*(T-1) & $3\times\sum^{d-1}_{i=1}[q\times(t-1)]^i$                                 & $3\times\sum^{d-1}_{i=1}[q\times(t-1)]^i$                                 \\
Avg. Calls per Instance                 & 1                  & 1    & 20     & 31.1          & 9.22                                                                      & 18.7                                                                      \\
Avg. Running Time per Instance (second) & 0.33               & 3.35 & 67.09  & 77.99         & 34.15                                                                     & 53.65                                                                     \\ \bottomrule
\end{tabular}
}
\caption{Evaluation of computational cost of different methods.} 
\label{tab:computational_cost}
\end{table*}

In Table~\ref{tab:computational_cost}, we provide the theoretical number of calls in CoT, SC-CoT, ToT and Socratic Questioning in 2 and 3 turns settings. We also provide the empirical results of the average number of calls per instance and average running time per instance in seconds for all methods. 
For SC-CoT, we fix the number of calls to 20 times on all the datasets based on the performance curve in~\cite{wang2022self}. 
In ToT, k represents the number of thoughts allowed to be generated per step, T represents the maximum number of steps and b represents the maximum number of states to keep at each step in BFS. Following~\cite{yao2023ToT}, we set k=5, T=3, and b=4. 
In Socratic Questioning, q represents the maximum number of raised sub-questions for a parent node.

As one can observe, Socratic Questioning with 2 turns and 3 turns achieves better efficiency compared to SC-CoT and ToT. The main reason is that, in the experimental datasets, most questions do not require a large amount of thinking steps to reach the correct answers. Socratic Questioning, adaptively raises sub-questions based on the complexity of the original question and arrives at the correct answer without reaching the theoretical maximum number of turns or depth. In contrast, both SC-COT and ToT employ fixed settings for the number of thoughts generated per step. For relatively straightforward questions, these fixed settings introduce high computational overhead, making the algorithms less efficient in these questions.

\section{Experimental Results on Other QA and Math Datasets}
Table~\ref{tab:extra_result} provides the performance of our method and two strong baselines on GSM8K and StrategyQA datasets. As one can observe, our method has significant performance improvement compared to baselines. We use ChatGPT with temperature 0.7 for all methods. For SC-CoT, we sample 20 reasoning paths.
\begin{table}[]
\centering
\resizebox{0.45\textwidth}{!}{%
\begin{tabular}{l|c|c}
\toprule
                     & \textbf{GSM8K} & \textbf{StrategyQA} \\ \midrule
CoT                  & 79.0           & 59.7                \\
SC-CoT               & 86.0           & 63.0                \\
Socratic-Questioning & \textbf{89.33}          & \textbf{65.33}               \\ \bottomrule
\end{tabular}
}
\caption{Accuracy (\%) on GSM8K and StrategyQA using Exact Match. The best performance is highlighted in \textbf{bold}.} 
\label{tab:extra_result}
\end{table}

We tried our best to reproduce the results of CoT and SC-CoT reported in~\cite{wang2022self} on StrategyQA. Following (Wang et al., 2022), we use the question-only set from BIG-bench collaboration (2021) and use the exact same prompt template and in-context examples in SC-CoT. However, we cannot reproduce the results on StrategyQA in~\cite{geva-etal-2021-aristotle} since Code-davinci-002 and Code-davinci-001 are no longer publicly available. In addition, our results of ChatGPT on StrategyQA also agree with more recent studies in~\cite{qin2023chatgpt}.

\section{Experiment Results based on GPT-4}
To showcase the generalizability of our approach, we have run CoT and Socratic Questioning on MMLU Chemistry and LogiQA based on GPT-4. The experimental results show that our Socratic Questioning approach still significantly outperforms CoT.
\begin{table}[]
\centering
\resizebox{0.45\textwidth}{!}{%
\begin{tabular}{l|c|c}
\toprule
                     & \textbf{MMLU Chemistry} & \textbf{LogiQA} \\ \midrule
CoT                  & 80.2           & 70.3                \\
Socratic-Questioning & \textbf{85.73}          & \textbf{75.3}               \\ \bottomrule
\end{tabular}
}
\caption{Accuracy (\%) of GPT-4 based approaches using Exact Match. The best performance is highlighted in \textbf{bold}.} 
\label{tab:extra_result_gpt4}
\end{table}

\section{Prmopt Templates}
\label{app:prompt}
 To make our method generalize to other reasoning domains, we carefully design in-context demonstrations to guide the LLM to generate more basic sub-questions in an efficient manner. More concretely, to create high-quality sub-questions in the in-context examples, we take the human reasoning process and domain knowledge into account and carefully annotate the sub-questions by ensuring that they are more basic questions compared to the original question and their solutions can contribute to the reasoning process of the original questions. For examples of sub-questions, please refer to Figure~\ref{fig:promt_qg}. Based on our experiments in math, physics, chemistry and VQA domains, we argue that with a few examples (5 in all our experiments) Socratic-Questioning can generalize to a new domain.

Following~\cite{Kadavath2022LLM_almost_know}, we ask the LLM itself to output a confidence level, “high”, “middle”, or “low”, towards its answer. In the in-context demonstrations, we label the correct answers with supportive hints in the context as “high” confidence, label the correct answers without supportive hints as “middle” confidence, and label incorrect answers as “low” confidence. In this way, we can guide the model to align its confidence to the correctness of the predicted answers. Our algorithm will continue raising sub-questions if the estimated confidence is not “high”. Please refer to Figure~\ref{fig:promt_qa}  for more examples.

\subsection{Prompts template of QA module}
\label{app:promt_qa}
\begin{figure*}
\centering
    \includegraphics[width=0.9\textwidth]{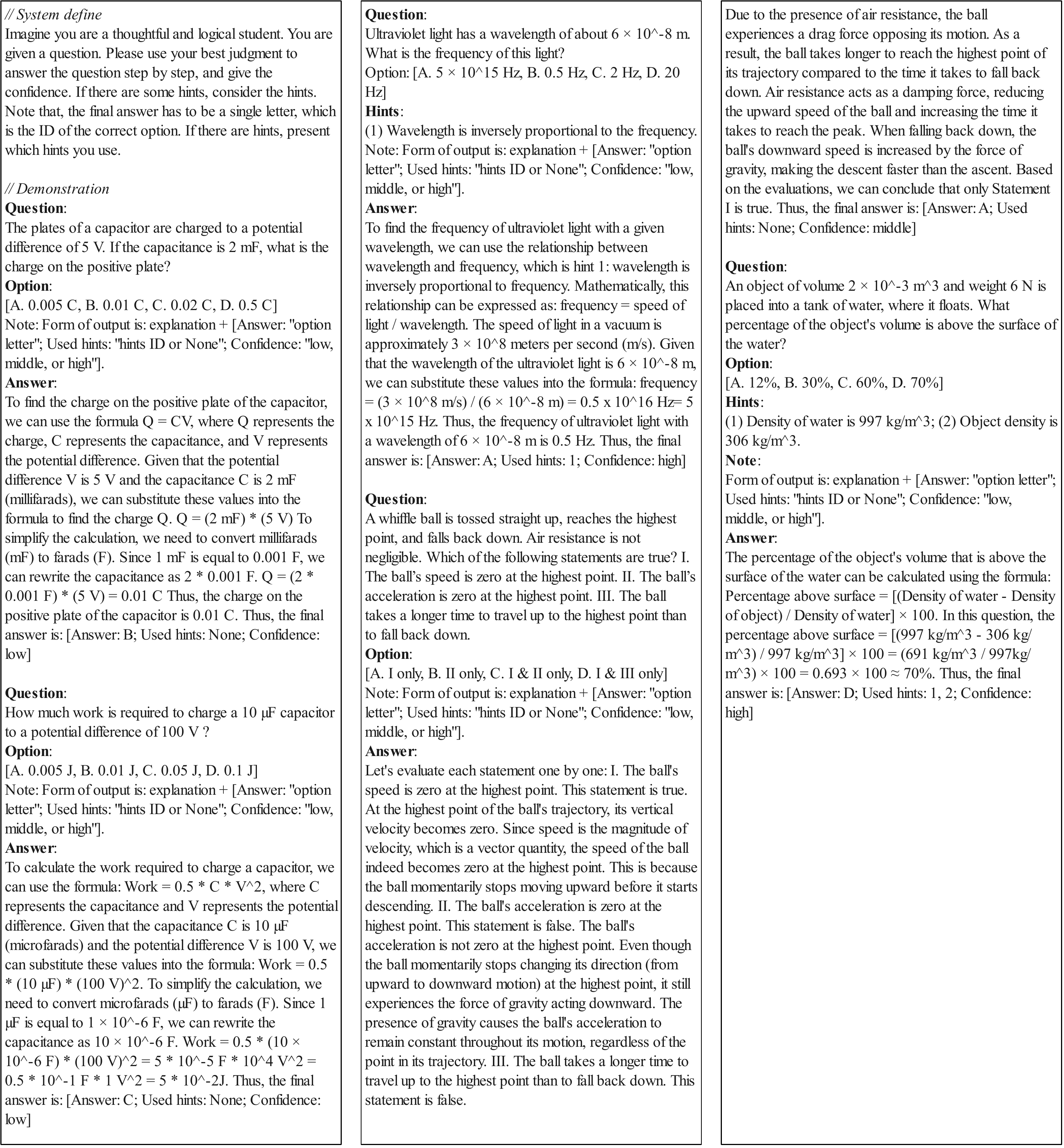}
    \caption{Prompt template of QA module.}
    \label{fig:promt_qa}
\end{figure*}

See Figure \ref{fig:promt_qa}.

\subsection{Prompts template of QG module}
\label{app:promt_qg}
\begin{figure*}
\centering
    \includegraphics[width=0.9\textwidth]{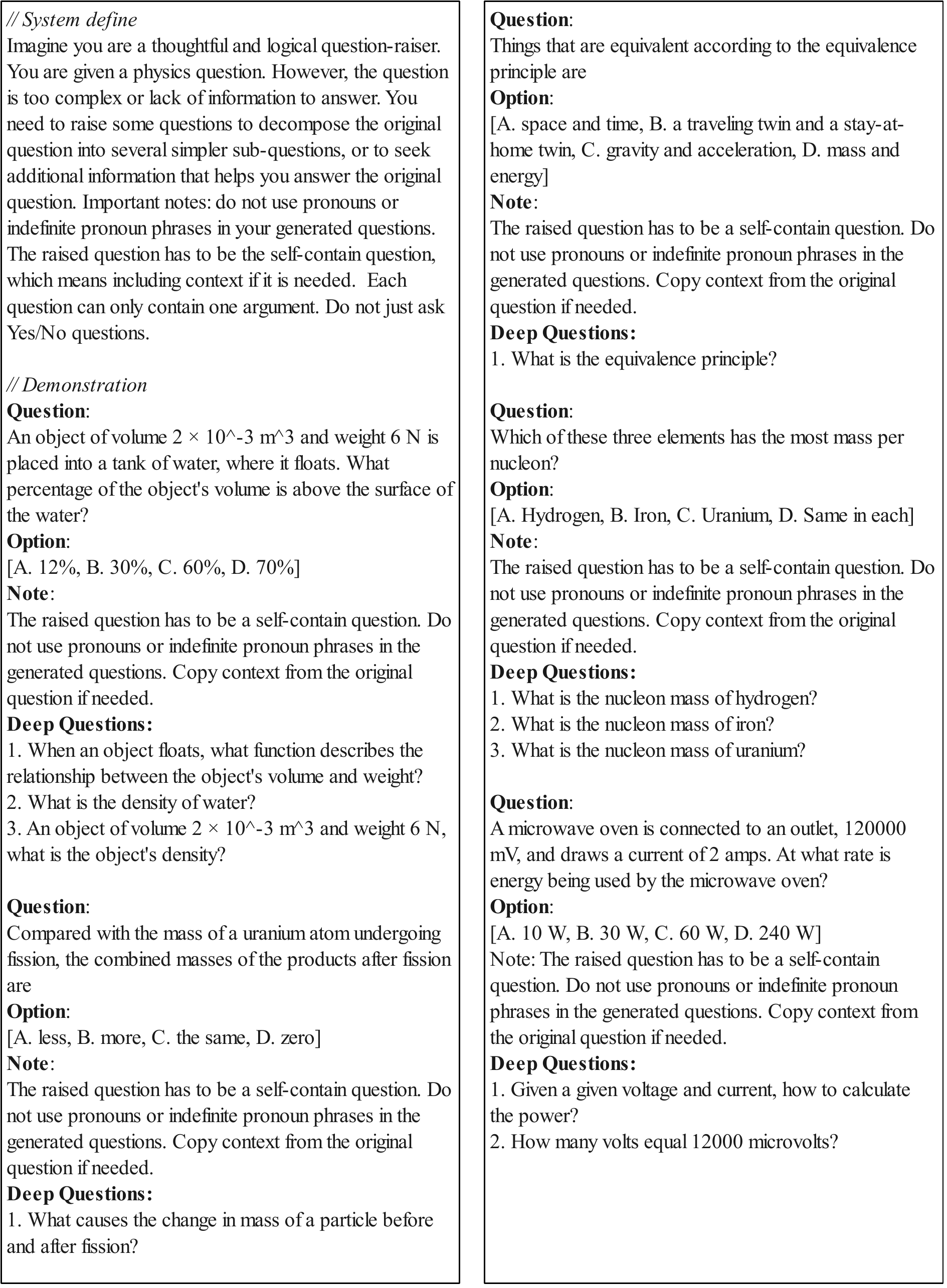}
    \caption{Prompt template of QG module.}
    \label{fig:promt_qg}
\end{figure*}

See Figure \ref{fig:promt_qg}.

\subsection{Prompts template of \FQG{}}
\label{app:fqg}
\begin{figure}
    \includegraphics[width=0.48\textwidth]{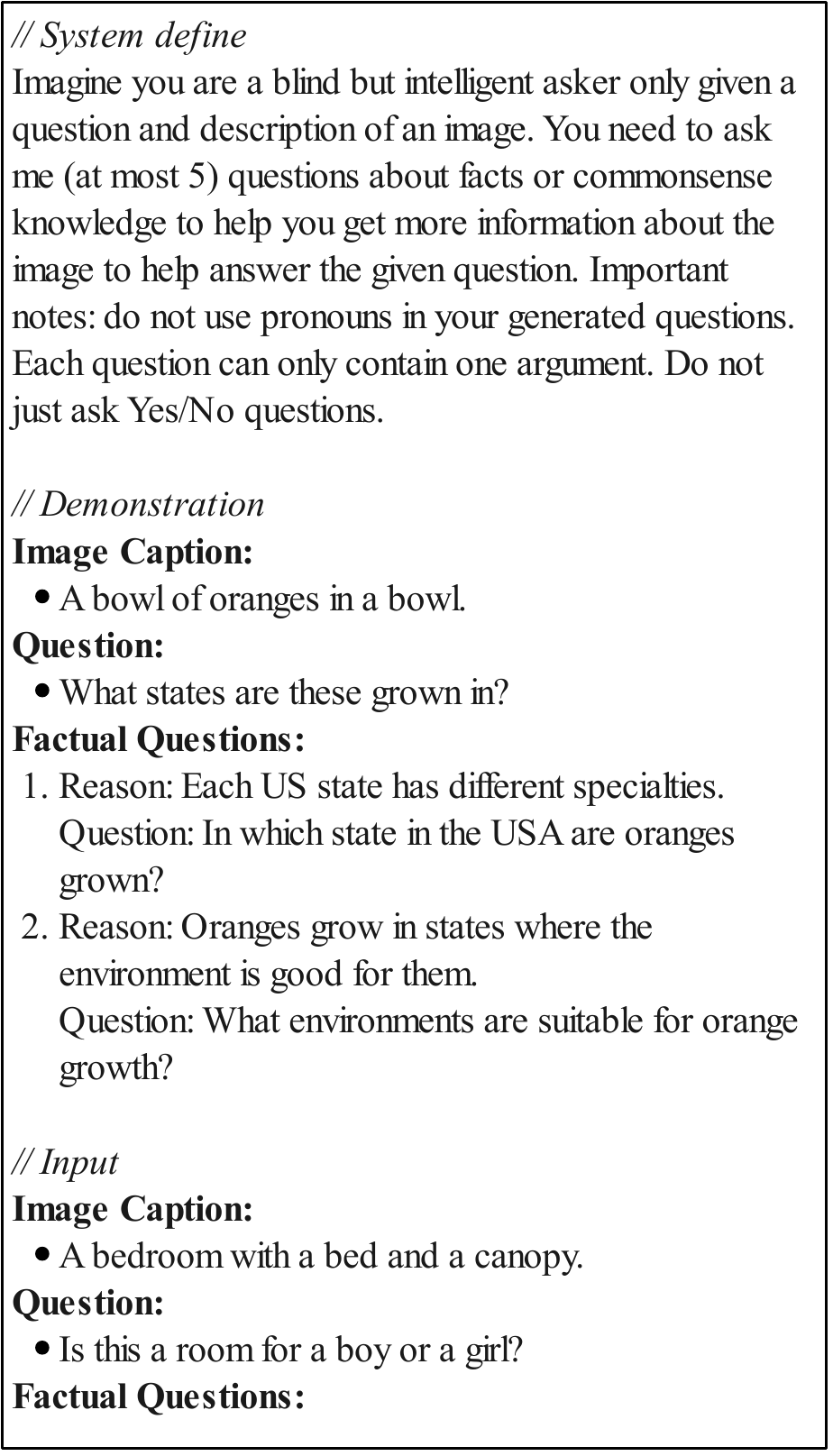}
    \caption{Prompt template of FQG module.}
    \label{fig:FQG}
\end{figure}

See Figure \ref{fig:FQG}.

\subsection{Prompts template of \FQA}
\label{app:kb}
\begin{figure}
    \includegraphics[width=0.48\textwidth]{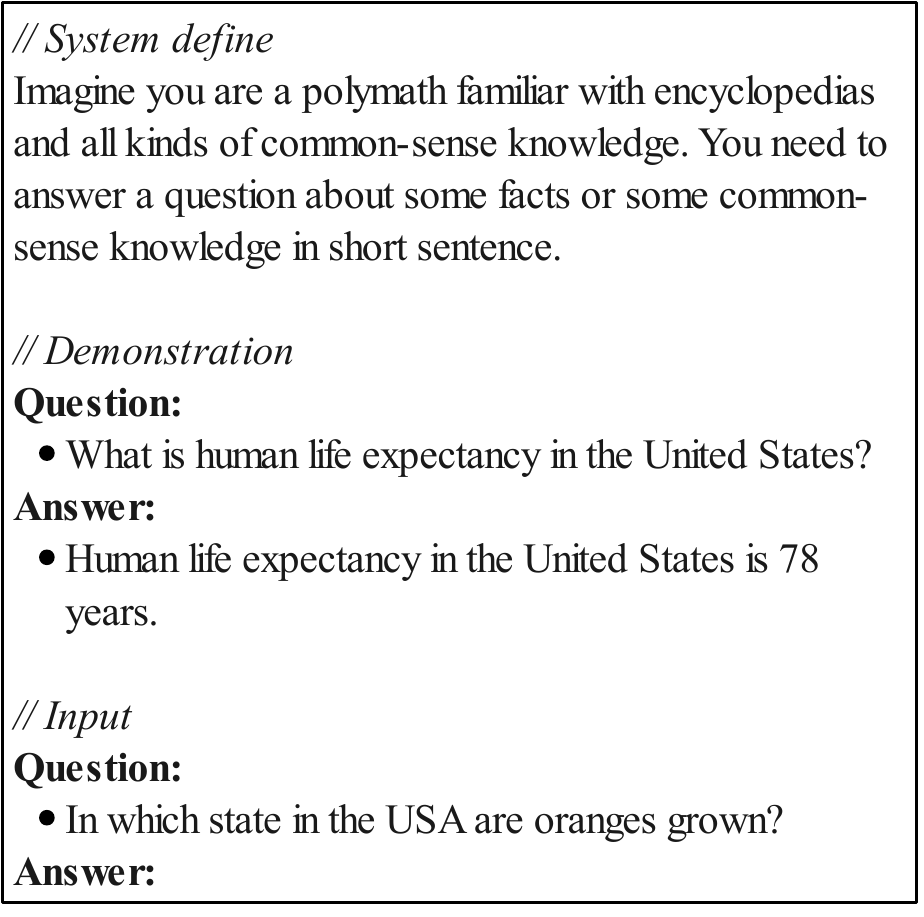}
    \caption{Prompt template of \FQA~module.}
    \label{fig:KB}
\end{figure}

See Figure \ref{fig:KB}.

\subsection{Prompts template of \VQG{}}
\label{app:vpg}
\begin{figure}
    \includegraphics[width=0.48\textwidth]{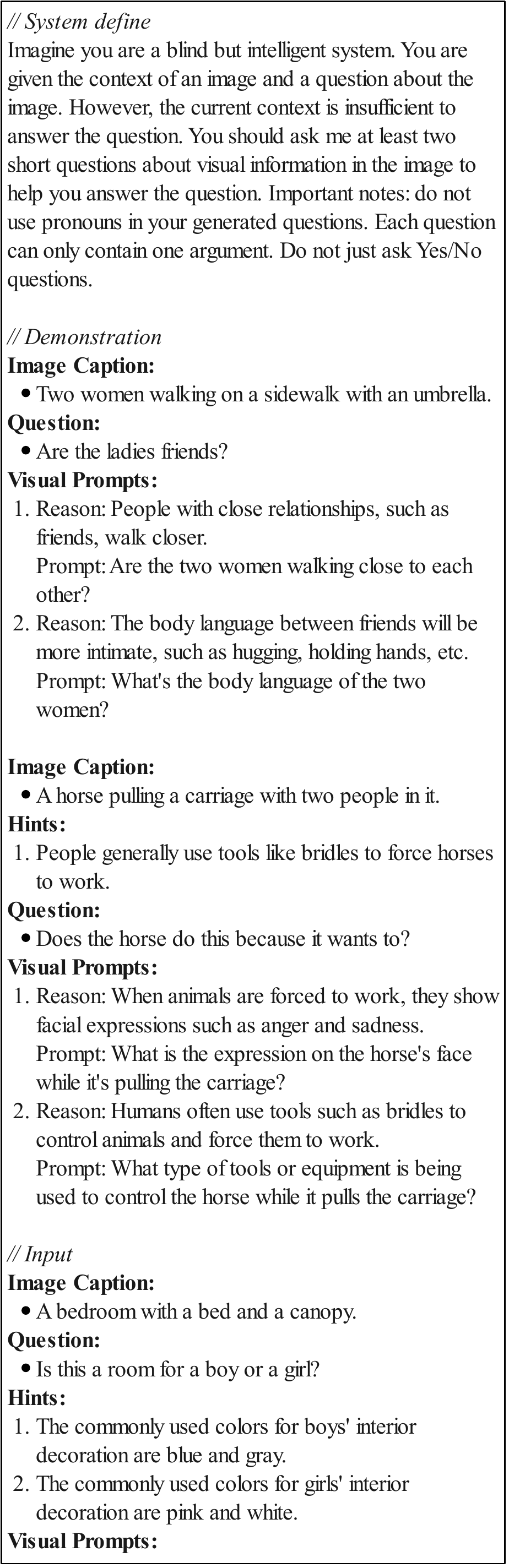}
    \caption{Prompt template of VQG module.}
    \label{fig:VPG}
\end{figure}

See Figure \ref{fig:VPG}.

\subsection{Prompts template of \VQA{}}
\label{app:tm}
\begin{figure}
    \includegraphics[width=0.48\textwidth]{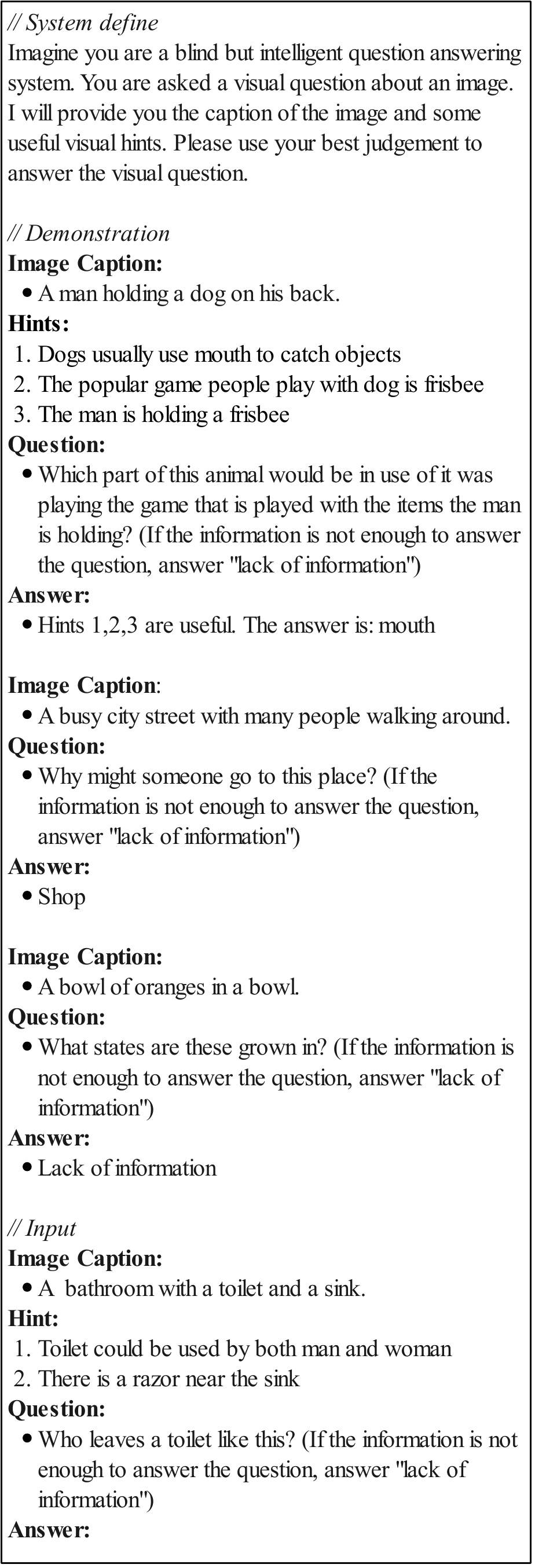}
    \caption{Prompt template of \VQA{} module.}
    \label{fig:TM}
\end{figure}

\begin{figure}
    \includegraphics[width=0.48\textwidth]{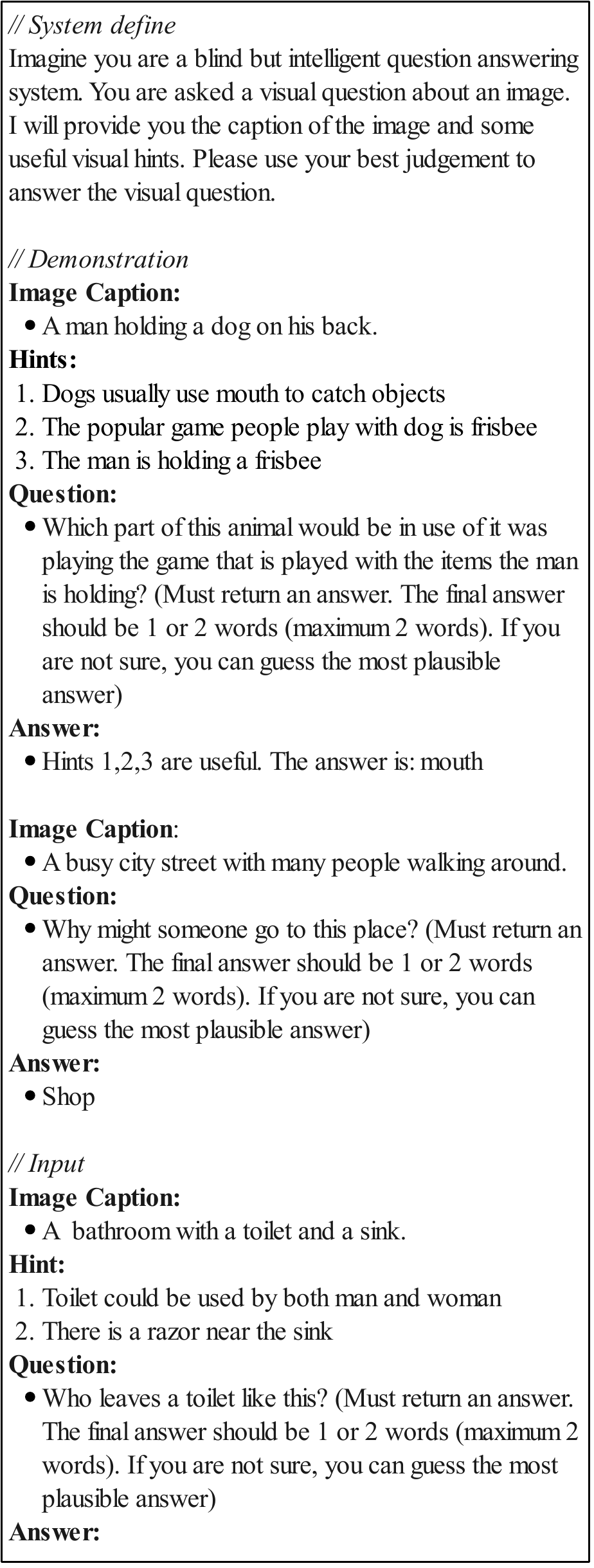}
    \caption{Prompt template of \VQA{} module (force answer).}
    \label{fig:TM_a}
\end{figure}

See Figure \ref{fig:TM} and \ref{fig:TM_a}.

\subsection{Prompts template of \QAhint{}}
\label{app:am}
\begin{figure}
    \includegraphics[width=0.48\textwidth]{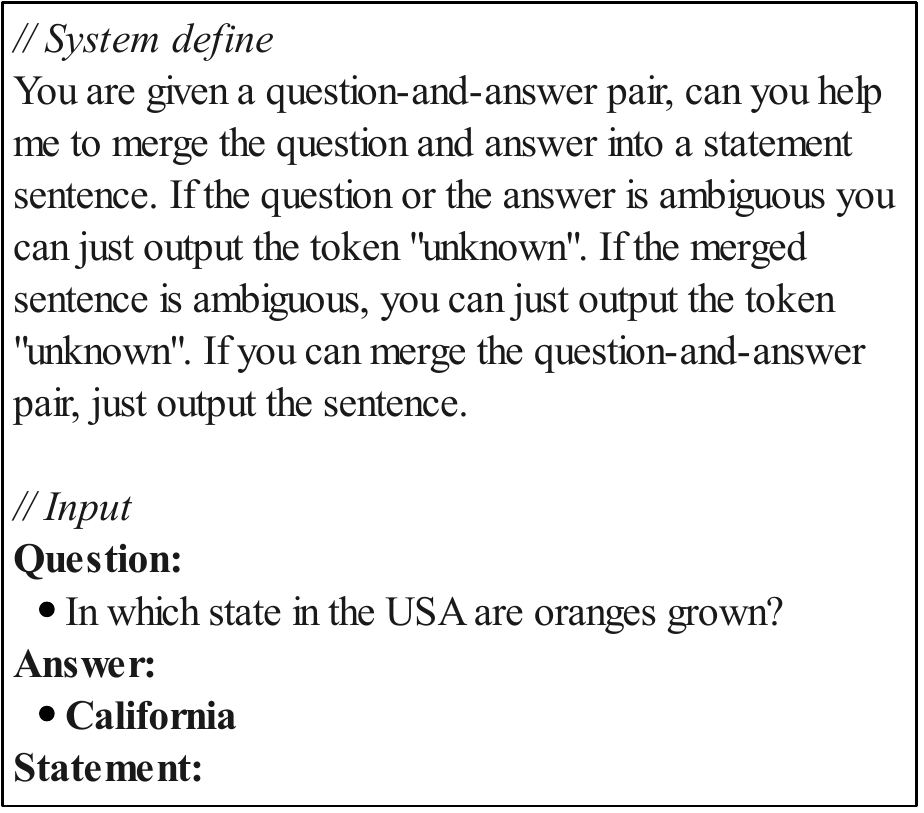}
    \caption{Prompt template of \QAhint{} module.}
    \label{fig:AM}
\end{figure}

See Figure \ref{fig:AM}.

\subsection{Prompt for chatGPT for semantic-based accuracy evaluation}
\label{app:grade}
\begin{figure}
    \includegraphics[width=0.48\textwidth]{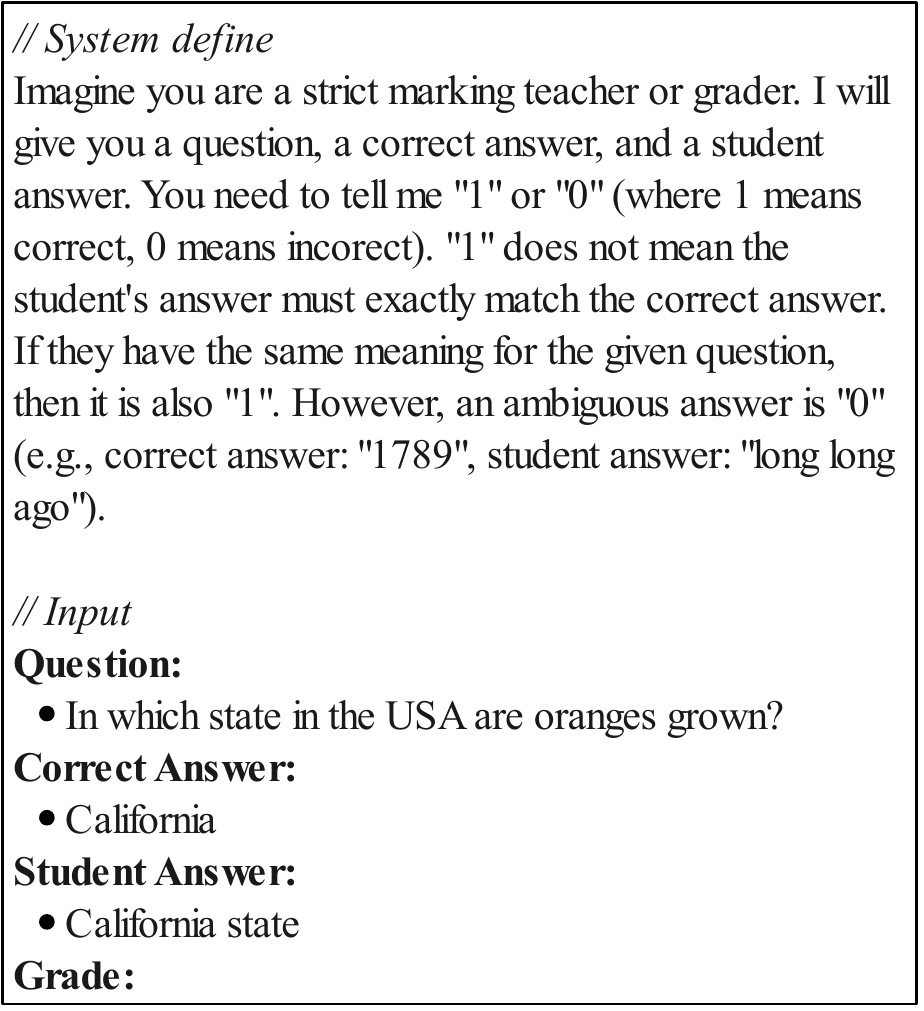}
    \caption{Prompt template for chatGPT of semantic-based evaluation.}
    \label{fig:grade}
\end{figure}

See Figure \ref{fig:grade}.


\end{document}